\documentclass[10pt,twocolumn,letterpaper]{article}

\usepackage[pagenumbers]{cvpr} %

\newcommand{\new}[1]{{\color{black}#1}}

\newcommand{\methodName}{SimLingo}
\newcommand\methodBase{\methodName{}-BASE}
\newcommand{\llm}{large language model}
\newcommand{\vlm}{vision language model}
\newcommand{\dream}{Action Dreaming}

\newcommand{\be}{\mathbf{e}}

\newcommand{\bi}{\mathbf{i}}\newcommand{\bI}{\mathbf{I}}

\newcommand{\bo}{\mathbf{o}}
\newcommand{\bp}{\mathbf{p}}
\newcommand{\bq}{\mathbf{q}}

\newcommand{\bw}{\mathbf{w}}

\newcommand{\nR}{\mathbb{R}}

\makeatletter
\DeclareRobustCommand\onedot{\futurelet\@let@token\@onedot}
\def\@onedot{\ifx\@let@token.\else.\null\fi\xspace}

\makeatother

\newcommand{\boldparagraph}[1]{\vspace{0.0cm}\noindent{\bf #1.}}

\definecolor{darkgreen}{rgb}{0,0.7,0}
\definecolor{darkyellow}{rgb}{0.8,0.8,0}
\definecolor{bittersweet}{rgb}{1.0, 0.44, 0.37}
\definecolor{amber}{rgb}{1.0, 0.49, 0.0}
\definecolor{lgray}{rgb}{0.83,0.83,0.83}

\definecolor{color_unlabled}{rgb}{0.0,0.0,0.0}
\definecolor{color_vehicle}{rgb}{0.0,0.0,0.56}
\definecolor{color_road}{rgb}{0.5,0.25,0.5}
\definecolor{color_redlight}{rgb}{1.0,0.0,0.0}
\definecolor{color_person}{rgb}{0.859,0.078,0.234}
\definecolor{color_roadline}{rgb}{0.613,0.914,0.195}
\definecolor{color_sidewalk}{rgb}{0.953,0.137,0.906}
\definecolor{teaser_red}{RGB}{222,112,97}

\definecolor{ellisred}{rgb}{0.87,0.44,0.38} %
\definecolor{ellisgreen}{rgb}{0.69,0.90,0.52} %
\definecolor{elliscyan}{rgb}{0.29,0.77,0.74} %
\definecolor{ellisorange}{rgb}{0.89,0.55,0.28} %
\definecolor{ellisblue}{rgb}{0.41,0.61,0.86} %

\definecolor{tuedgray}{RGB}{56,55,55}
\definecolor{tuelgray}{RGB}{246,246,246}
\definecolor{tuedblue}{RGB}{26,58,91}
\definecolor{tuelblue}{RGB}{133,203,210}
\definecolor{tueoblue}{RGB}{119,221,204}
\definecolor{tueogreen}{RGB}{119,221,159}
\definecolor{tuesgreen}{RGB}{186,213,72}
\definecolor{tueyellow}{RGB}{255,221,0}
\definecolor{tuered}{RGB}{234,75,46}

\mathchardef\mhyphen="2D

\usepackage{booktabs}       %
\usepackage{multirow}
\usepackage{graphicx}
\usepackage{colortbl}
\usepackage{textcomp}
\usepackage{enumitem}

\definecolor{cvprblue}{rgb}{0.21,0.49,0.74}
\usepackage[pagebackref,breaklinks,colorlinks,allcolors=cvprblue]{hyperref}

\def\paperID{1373} %
\def\confName{CVPR}
\def\confYear{2025}

\title{\methodName: Vision-Only Closed-Loop Autonomous Driving with Language-Action Alignment}

\author{
Katrin Renz$^{1,2,3\ast}$ \quad
Long Chen$^{1}$ \quad
Elahe Arani$^{1}$ \quad
Oleg Sinavski$^{1}$ \\
[2mm]
$^1$~Wayve\quad
$^2$~University of Tübingen \quad
$^3$~Tübingen AI Center \quad
}

\begin{document}

\twocolumn[{%
\renewcommand\twocolumn[1][]{#1}%
\vspace{-14.0mm}
\maketitle
\begin{center}
    \centering
    \captionsetup{type=figure}
    \includegraphics[width=0.8\textwidth]{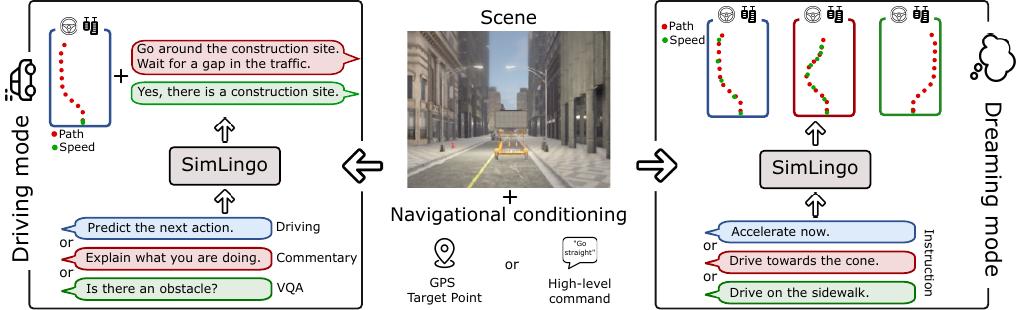}
    \captionof{figure}{\label{fig:teaser}
    \textbf{Overview}: \methodName{} is a vision-language-action model unifying the tasks of autonomous driving, vision-language understanding and language-action alignment. It is state of the art on the official CARLA Leaderboard 2.0 and Bench2Drive using only camera images. We introduce the task of \dream{}, a form of instruction following, to improve the alignment of language and action.
    }

\end{center}}]

{\let\thefootnote \relax \footnote{$^*$Work done while interning at Wayve.}}
{\let\thefootnote \relax \footnote{Challenge tech report (preliminary version):}
{\let\thefootnote \relax \footnote{\href{https://arxiv.org/abs/2406.10165}{ \texttt{https://arxiv.org/abs/2406.10165}}}}

\begin{abstract}
Integrating \llm{}s (LLMs) into autonomous driving has attracted significant attention with the hope of improving generalization and explainability. However, existing methods often focus on either driving or vision-language understanding but 
achieving both high driving performance and extensive language understanding remains challenging. 
In addition, the dominant approach to tackle vision-language understanding is using visual question answering.
However, for autonomous driving, this is only useful if it is aligned with the action space. Otherwise, the model's answers could be inconsistent with its behavior.
Therefore, we propose a model that can handle three different tasks: (1) closed-loop driving,  (2) vision-language understanding, and (3) language-action alignment. 
Our model \methodName{} is based on a \vlm{} (VLM) and works 
using only camera, excluding expensive sensors like LiDAR.
\methodName{} obtains state-of-the-art performance on the widely used CARLA simulator \new{on the Bench2Drive benchmark and is the winning entry at the CARLA challenge 2024}. Additionally, we achieve strong results in a wide variety of language-related tasks while maintaining high driving performance. 
\end{abstract}
\vspace{-0.3cm}
\section{Introduction}
\label{sec:intro}
In recent years \vlm{}s (VLMs)~\cite{liu2023llava, liu2024llavanext, chen2024internvl, Qwen2VL} demonstrated their capacity for having a broad knowledge about the world and being able to generalize to unseen prompts. 
Incorporating these capabilities into autonomous driving systems has the potential to enhance robustness in diverse, rare scenarios and enable smoother human-machine interaction.
The field of vision-driven
robotics has begun to leverage pre-trained \vlm s (VLMs) to enhance performance 
as models are better in understanding and adapting to unfamiliar environments and in following natural language commands~\cite{zitkovich2023rt, black2024pi_0, kim24openvla}. \\
In end-to-end autonomous driving the introduction of VLMs into the driving models also gained traction~\cite{Sima2024ECCV, lingo2, hwang2024emma}.
VLMs are used to improve planning performance~\cite{shao2023lmdrive, Zhang2024CVPR} or enable vision-language capabilities in the form of visual question answering (VQA) tailored around driving scenes~\cite{qian2023nuscenesqa, Marcu2024ECCV}. 
VQA is a promising way to test a model's ability to comprehend the scene and understand the significance of objects for their behavior. However, when evaluating this understanding only in language space, it can be completely disentangled from the actual driving decision (e.g. the model claims to see a red traffic light but the actions indicate to accelerate). We argue that only \textit{aligned} language interactions - where the model’s actions adapt in response to language cues - can provide causal evidence of true language understanding.
\new{
To address this challenge of aligning language understanding with driving actions, we propose \textit{\dream{}} - a new dataset to improve alignment and a benchmark for evaluating how language inputs influence actions without executing dangerous instructions. 
When generating instruction-action pairs post-hoc on existing expert data, the action can be inferred from visual cues alone. Therefore, the model does not need to pay attention to the language instruction. This results in a misalignment between language and action.
To address this, we propose a novel data-collection technique where we simulate multiple possible futures for a given state. We train and test on a set of diverse instruction-action pairs for the same visual context, ensuring that the model listens to the instruction. \\
The second main focus is the driving performance. 
Many works that combine VLMs with driving evaluate their performance in either overly simplistic environments (e.g., HighwayEnv)~\cite{Wen2024ICLR, Daocheng2023WACVW, Cui2023ARXIVc, Ma2024CVPR} or in an open-loop setting (e.g., NuScenes planning)~\cite{Sima2024ECCV, hwang2024emma, Tian2024ARXIVb, Mao2024ARXIV}. We rigorously test our models on a challenging closed-loop benchmark.
}

\boldparagraph{Contributions}
(1) A VLM-based driving model that achieves state-of-the-art driving performance on the official CARLA Leaderboard 2.0 and the local benchmark Bench2Drive in the CARLA simulator. (2) A new task (\dream), which comes with a methodology to collect instruction-action pairs and a benchmark to evaluate the connection of language and action understanding without having to execute unsafe actions. (3) A generalist model that achieves not only good driving performance but also includes several language related tasks in the same model.
\section{Related Work}
\boldparagraph{End-to-end autonomous driving}
End-to-end training based on Imitation Learning (IL)~\cite{Chen2022CVPRa, Wu2022NeurIPS, Shao2022CORL, jaeger2023hidden} is the dominant approach for
state-of-the-art methods on the CARLA Leaderboard (LB) 1.0~\cite{Leaderboard2020}. With the introduction of the CARLA Leaderboard 2.0~\cite{Leaderboard2024} the driving task became fundamentally harder. This is shown by applying one of the leading methods of LB 1.0 (TransFuser) zero-shot to the new LB 2.0, which obtains a huge drop in driving score (66.32 to 0.58).
Most state-of-the-art end-to-end methods incorporate auxiliary outputs and rely on multiple sensors like camera combined with LiDAR~\cite{jaeger2023hidden, Chitta2023PAMI, Shao2023CVPR, Shao2022CORL, Chen2022CVPRa}. However, there is a growing interest in deploying camera-first models, as shown by industry players \cite{Tesla2014,lingo2} and academic work like DriveCoT~\cite{Wang2024ARXIVdrivecot} and TCP~\cite{Wu2022NeurIPS} show competitive results with using only camera images. We follow this approach and exclude expensive sensors like LIDAR. 

\boldparagraph{Language models for driving} 
Most state-of-the-art driving models introduce domain-specific architectures~\cite{hu2023_uniad,Chitta2023PAMI, jaeger2023hidden}.
However, with the recent progress with \llm{}s and \vlm s, those generalist architectures have been integrated into driving systems. Multi-modal LLM-based driving frameworks such as LLM-Driver \cite{Chen2023ICRA}, DriveGPT4 \cite{xu2023drivegpt4}, and DriveLM \cite{Sima2024ECCV} utilize foundation models with inputs from different modalities for driving.
GPT-Driver \cite{mao2023gptdriver} and LanguageMPC \cite{sha2023languagempc} fine-tune ChatGPT as a motion planner using text. Knowledge-driven approaches \cite{Wen2024ICLR, Daocheng2023WACVW} are also adopted to make decisions based on common-sense knowledge. However, most of these works are evaluated primarily through qualitative analysis in open-loop settings like NuScenes or in simplified driving environments like the HighwayEnv~\cite{Wen2024ICLR, Daocheng2023WACVW, Cui2023ARXIVc, Ma2024CVPR}.
FED~\cite{Zhang2024CVPR} uses a VLM for closed-loop driving by using language to provide feedback during training. 
LMDrive~\cite{shao2023lmdrive} proposes an end-to-end closed-loop driving model that can follow a limited set of human instructions. \new{
However, when evaluated on commonly used driving benchmarks, it is far behind state-of-the-art driving models~\cite{li2025comprehensive}.
}
DriveMLM~\cite{Wang2023ARXIVb} also includes instruction following \new{but unlike ours it uses an off-the-shelf motion planner and is not trained end-to-end.}

Several studies have integrated language understanding into autonomous driving systems through Visual Question Answering (VQA). However, many approaches either limit their evaluation to open-loop settings~\cite{Sima2024ECCV, Lu2024ARXIV, hwang2024emma} or do not involve action prediction at all~\cite{Marcu2024ECCV, qian2023nuscenesqa, Ma2024ARXIV}. \new{
So far, open-loop models are far behind when tested on closed-loop benchmarks~\cite{hu2022stp3, jiang2023vad, Mao2024ARXIV}.
There is no evidence that open-loop results transfer to closed-loop, making ablations and claims unreliable~\cite{jaeger2024github}. This highlights that bridging this gap requires major changes in training, architecture, control, and data composition.
}
Additionally, there is a fundamental lack of systematic evaluation of instruction-following abilities, particularly when looking at the alignment of instruction and action.

\section{Method}
\label{sec:method}
\begin{figure*}[t]
\begin{center}
   \includegraphics[width=0.7\linewidth]{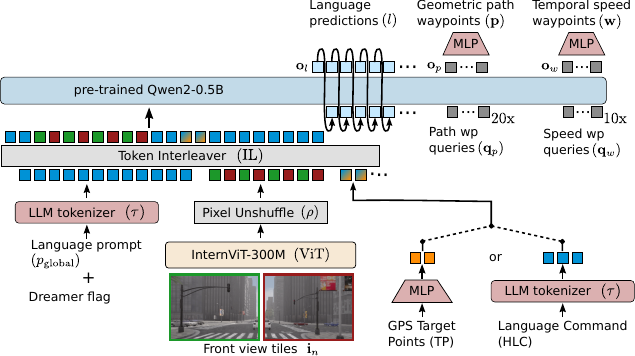}
\end{center}
\vspace{-0.4cm}
\caption{\textbf{\methodName{} architecture.} We encode the image, navigational conditioning and the language prompt. To encode high-resolution images, we split them into tiles, and encode each independently to reuse the pre-trained image encoder pre-trained on 448x448 resolution. All embeddings get processed by an LLM which we finetune with LoRA to predict language and actions. The action output utilizes a disentangled representation with both temporal speed waypoints and geometric path waypoints for improved lateral control.}
\label{fig:architecture}
\vspace{-0.3cm}
\end{figure*}

\subsection{Task overview}\label{sec:task}
\boldparagraph{Driving}
The driving objective is to reach a specified target location on a given map while passing predetermined intermediate locations in the CARLA \cite{CarlaDosovitsky} autonomous driving simulator. The map includes diverse environments such as highways, urban streets, residential areas, and rural settings, all of which must be navigated under various weather conditions, including clear daylight, sunset, rain, fog, and nighttime scenarios.
Along the way, the agent must manage dense occurrences of various complex scenarios such as encountering jaywalkers, navigating parking exits, executing unprotected turns, merging into ongoing traffic, passing construction sites, or avoiding vehicles with an opened door.
\boldparagraph{Vision-Language Understanding}
In addition, the model needs to solve a diverse set of language tasks, including describing the current decision and action in natural language (Commentary) and answering questions about the simulated driving scene (VQA). 
The task of Commentary answers the question ``What should the ego vehicle do next and why?". The answer refers to the path (e.g. changing lanes, going around objects) and the speed together with the reason for both. 
Additionally, we run Commentary by default in our final model during inference. This means the final driving model acts in a Chain-of-Thought setting, where first the action and reason are predicted in language space, and then the action is predicted, conditioned on the generated commentary.
\boldparagraph{\dream{}}
The third task combines vision-language understanding with the ability to align it with the action space. 
We task the model with a wide variety of instructions including change of speed, lane changes, driving towards specific objects, or other navigational changes. This includes common instructions that can occur during normal driving but also out-of-distribution instructions (e.g., crashing into objects) that should never be executed in the real world but are indicative of the model's understanding of diverse instructions. This tests whether the model's language capabilities are aligned with the action space. Since we do not want to execute those actions we call this the \textit{\dream{}} mode. We provide the model with a language instruction and the model needs to predict the corresponding action.

\subsection{Datasets}\label{sec:data}
\boldparagraph{Driving} We utilize the privileged rule-based expert \textit{PDM-lite} \cite{Beißwenger2024TECH, Sima2024ECCV} to collect our driving dataset in the CARLA simulator. 
We use three sets of routes with scenarios for data collection: (1) training routes of TransFuser~\cite{Chitta2023PAMI} of Town 1-10 which we converted to support Leaderboard 2.0. (2) We divide the official CARLA LB 2.0 routes of Town 12 and Town 13 into shorter segments centered around a single scenario to reduce trivial data (e.g., driving straight without any hazardous events) and to simplify data management as proposed by~\cite{Chitta2023PAMI, li2024think2drive}. (3) With the introduction of LB 2.0, the maximum distance between navigational target points increased from 50 to 200 meters. The routes with one scenario often fall within this distance, causing a distribution shift, as the next target point is at the end of the route (i.e., closer than 200m) rather than after 200 meters. Consequently, we employ a set of longer routes, featuring three scenarios per route.
To ensure balance of scenario types, we adjust the number of routes per scenario (i.e., upsampling routes with rare scenarios), apply random weather augmentation, and modify how far scenarios get spawned by ±10\%.
Overall, we collect 3.1 million samples at 4 fps.
\boldparagraph{Vision-Language Understanding} For the vision-language understanding data we apply the heuristics of DriveLM~\cite{Sima2024ECCV} to generate VQA labels for our driving dataset. We use Town 12 for training and Town 13 for evaluation.
For the commentary task, we generate new labels as this task is not part of the DriveLM labels.
Similarly, we use privileged information during data collection and the logic of the rule-based expert to generate explanations. More precisely, we use the leading object obtained from the experts' Intelligent Driver Model (IDM)~\cite{Treiber_2000} as well as information about changing the path to swerve around objects. In addition, we use heuristics based on the ego waypoints to distinguish between driving intentions like starting from stop or keep driving at the same speed. 
\boldparagraph{\dream{}}
To align language with the action space we need data with paired instruction and action.
Relying solely on the actual actions executed by the expert for generating instructions post-hoc, results in labels that can be entirely inferred from the available visual information, thus making it unnecessary to process the language instructions. To address this limitation, generating multiple alternative instructions and actions for the same visual context forces the model to listen to the instruction.
For the original dataset, we store a non-exhaustive portion of the simulator's state, such as the positions of other vehicles and their speeds. We utilize the "world-on-rails" assumption~\cite{Chen2021ICCVa} (i.e., all other dynamic agents are treated as being on fixed paths and are unresponsive to environmental changes, akin to replaying the original simulation) for other actor and 
approximate the ego vehicle’s dynamics using a kinematic bicycle model.
This allows us to simulate different possible future trajectories of the ego vehicle and perform collision checks with other agents without needing to execute the actions.
This dataset incorporates several types of new trajectories, such as lane changes (including movements onto sidewalks or parking lanes), speed adjustments, driving toward specific objects (e.g., traffic cones, signs, vehicles) potentially causing collisions, and crossing particular road markings like stop signs or stop lines. For each alternative instruction-action pair we also have a flag if this action would be safe to execute and a language label describing if the action can be executed or not together with the reason in case not. We use Town 12 for training and Town 13 for validation.
More details about the data collection and labelling procedure can be found in Appendix~\ref{sec:supp:datasets}.

\subsection{Architecture} \label{sec:arch}

\methodName{} is built on top of the InternVL-2 architecture~\cite{gao2024mini} with additional input modalities and output heads.
Fig.~\ref{fig:architecture} shows an overview of the \methodName{} architecture.

\boldparagraph{Input Representation}
\methodName{} uses a camera image to navigate the environment. We input the camera image $\bI \in \nR^{H \times W}$, navigational information in the form of either the next two GPS target points $\mathrm{TP}$ or a high-level language command $\mathrm{HLC}$ (e.g. "turn left"), and the ego vehicle's speed $v$.
Additionally, we input a task prompt with information about the current task. 
We differentiate between 4 task prompts $p_{\mathrm{task}}$: (1) Driving without language predictions: ``Predict the waypoints." (2) Commentary + Driving: ``What should the ego do next?" (3) VQA + Driving: ``Q: \textit{\textlangle question\textrangle}?" (4) \dream: ``\textlangle Dreamer flag\textrangle\textit{\textlangle instruction\textrangle}." For the \textlangle Dreamer flag\textrangle{} we differentiate between two modes: when activated the model should predict actions aligned with the instruction, when deactivated, the model needs to reject unsafe instructions.
All input information is part of the global LLM prompt $p_{\mathrm{global}}$, which has the following structure: ``\textit{\textlangle image features $\be_{I}$\textrangle \textbackslash n Current speed: \textlangle $v$\textrangle m/s. Command: \textlangle nav. features $\be_{nav}$\textrangle. \textlangle task prompt $p_{\mathrm{task}}$\textrangle.}''
All parts inside \textlangle\textrangle{} get replaced by the corresponding feature embeddings.

\boldparagraph{Output Representation} 
The output consists of two modalities: action and language. For any language prediction $l$ we use standard auto-regressive token prediction.
For the action representation, we use a disentangled representation with (1) temporal speed waypoints $\bw \in \nR^{N_w \times 2}$, with $N_w$ future coordinates with one coordinate every 0.25 seconds. This represents the location of the ego vehicle at a specific time in the future. Also, we predict (2) geometric path waypoints $\bp \in \nR^{N_p \times 2}$, with $N_p$ future coordinates with one coordinate every meter. This represents future positions of the ego vehicle independently of the time to reach them.
From the temporal waypoints $\bw$ we obtain a target speed and from the geometric waypoints $\bp$ a target angle. We then use two PID controllers to get the steering angle and acceleration.
We noticed that using only temporal waypoints for steering and speed led to steering problems, especially during turns or when swerving around obstacles. By using path waypoints, we achieve denser supervision, as we also predict the path when the vehicle is stationary, leading to improved steering behavior. 
To enable more efficient action prediction we predict all action embeddings in one forward pass instead of auto-regressively. For this, we input learnable query tokens $\bq_p$ and $\bq_w$. An MLP on top of the output features $[\bo_p, \bo_w]$ generates waypoint differences. The cumulative sum of these differences yields the final waypoints $\bp$ and $\bw$.\looseness=-1

\boldparagraph{Vision-Language-Action Model}
We use the InternVL2-1B from the Mini-InternVL family~\cite{gao2024mini} as our main architecture for \methodName{}. The InternVL2 family~\cite{gao2024mini,chen2024far, chen2024internvl} reaches state-of-the-art performance while providing the widest range of available model sizes starting with a 1B parameter model.
The InternVL2-1B model consists of the vision encoder InternViT-300M-448px ($\mathrm{ViT}$) and Qwen2-0.5B-Instruct~\cite{qwen2} as the language model ($\mathrm{LLM}$). \\
\textit{Variable high-resolution image encoding.} Processing high-resolution images is crucial for detailed image understanding. Especially for autonomous driving important information, such as traffic lights at large intersections, may only be visible in a few pixels. Most VLMs are based on CLIP pre-trained vision encoder which are pre-trained on a resolution of 336x336 (some 448x448).
To be able to process dynamic and higher resolutions we split the input image $\bI \in \nR^{H \times W}$  into $N_{I}$ 448x448 pixel tiles $\bi_n \in \nR^{448 \times 448}$  and extract features for each tile independently. This method is commonly used in the latest VLMs~\cite{chen2024internvl, liu2024llavanext}.
To reduce computational overhead due to the nature of the quadratic complexity of the LLM, InternVL2 uses the pixel unshuffle technique~\cite{Shi2016CVPRa} ($\rho$) to downsample the number of tokens by a factor of 4. 
Each 448 × 448 tile is then represented by 256 visual tokens. 
We obtain visual features $$\be_I = \rho([\mathrm{ViT}(\bi_n)]_{n=0}^{N_{I}}) \quad \in \nR^{(N_{I}*256) \times D}$$ with embedding size $D$. In this work we use $N_{I}=2$ resulting in 512 visual tokens.\\
\textit{Driving specific inputs.} 
The navigational information is presented either as target points in the form of GPS locations (TP) or as a language command (HLC). When using the target points, we encode the locations with an MLP to obtain two navigational embeddings $\be_{nav} \in \nR^{2 \times D}$. 
In the case of using the language command we use the standard tokenizer of the LLM to obtain the embeddings $\be_{nav} \in \nR^{N_{\mathrm{HLC}} \times D}$, with $N_{\mathrm{HLC}}$ equals the number of tokens representing the navigational command. 
During training, we randomly switch between the two input modalities. 
We use the speed $v$ in natural language as part of the global LLM prompt.\looseness=-1 \\
\textit{Language input.}
For all language parts, we use the original tokenizer of the LLM to obtain the token embeddings $\be_L$. \\
\textit{Token interleaver.} 
After encoding each input modality the token interleaver ($\mathrm{IL}$) replaces the placeholder tokens \textlangle\textrangle{} of $p_{\mathrm{global}}$ with the corresponding embeddings, to obtain the input embedding sequence:
$$\be_{\mathrm{LLM}} = \mathrm{IL}(\be_L, \be_I, \be_{nav}).$$ This interleaved token sequence is then input to the pretrained LLM. \\
\textit{Large language model.}
Given the input embeddings and the action queries the LLM generates language and action output features
$$
[\bo_l, \bo_p, \bo_w] = \mathrm{LLM}([\be_{\mathrm{LLM}}, \bq_w, \bq_p]) \quad .
$$ 
First, it auto-regressively generates the language predictions, which represent the answer to the task prompt. Then in one additional forward pass, it generates the actions consisting of path and waypoints.
We use the smooth-L1 loss on the path and waypoints and cross-entropy loss on the predicted language tokens.

\subsection{Training} \label{sec:train}
\boldparagraph{Dataset mixtures}
We differentiate between two action labels: (1) expert trajectories and (2) dream trajectories (safe and unsafe ones). During training, we sample 50/50 from each.
For the expert trajectories, we do one of three options for additional language supervision: (1) VQA, (2) Commentary, and (3) no language. We sample 50\% VQA prediction, 35\% commentary prediction, 7.5\% we provide the commentary in the prompt, and 7.5\% we do not supervise any language.
For the dream trajectories, we train 50\% with the \textit{Dreamer flag} activated and 50\% with it being deactivated, where the model needs to decide if the instruction is safe to execute and reject unsafe ones.
\boldparagraph{Data buckets}
The majority of driving involves straight, uneventful segments. To maximize the collection of interesting scenarios, we focus the data collection around a diverse range of challenging situations. However, some ratio of easy and uneventful data is inevitable. To address this issue, we create data buckets containing specific interesting samples and assign a probability to each bucket. During training, we sample from these buckets instead of the entire dataset.
This approach reduces the number of samples per epoch to 650,000. Details about the bucket types are in the Appendix~\ref{sec:supp:buckets}.

\section{Experiments}
\label{sec:experiments}
In this section, we start with an overview of the used benchmarks and metrics (Section~\ref{sec:bench}) and the implementation details (Section~\ref{sec:impl}). We then demonstrate our results (Section~\ref{sec:results}) quantitatively and qualitatively. 
\subsection{Benchmarks and Metrics}\label{sec:bench}
For a detailed description of the metrics we refer to Appendix~\ref{sec:supp:metric}.
\boldparagraph{Leaderboard 2.0} We use the official test server of the CARLA simulator with secret routes under different weather conditions~\cite{Leaderboard2024}. We report the official CARLA metrics, Driving Score (DS), Route Completion (RC), and Infraction Score (IS).
The Driving Score is calculated in a way that models in the current performance range get punished for completing more of the route due to a non-linear decrease of the score due to infractions.
A more detailed discussion about its flaws is in the Appendix~\ref{sec:supp:metric}. \\
\boldparagraph{Bench2Drive} For local evaluation we use the Bench2Drive Benchmark~\cite{jia2024bench} based on CARLA version 0.9.15 which uses 220 routes with approximately 150m on Town01 - Town15 and different weathers. 
We use the official metrics, which include driving score, success rate, efficiency, and comfortness. Those shorter routes make it easier to obtain higher driving scores making the range not comparable with the Leaderboard numbers.\\
\boldparagraph{DriveLM-hard (VQA) and Commentary} We use DriveLM~\cite{Sima2024ECCV} to generate VQA data and our annotation scheme described in Section~\ref{sec:data} to collect additional Commentary labels for Town 13 as validation labels which we withhold during training. 
Differently to DriveLM, we construct a harder validation set for which we uniformly sample 10 examples per answer type instead of randomly sampling from the whole set and therefore highly undersample rare answers. Our VQA validation set contains 330 different answer types and the Commentary one contains 190. We use the \textit{GPT} and \textit{SPICE} metric.
\\
\boldparagraph{\dream{}} To assess not only the ability to understand scene-specific knowledge in language space but also to connect this to the action space we introduce the \textit{\dream{}} evaluation. The model gets a language input in the form of an instruction and needs to imagine how the corresponding actions would look like. We use Town 13 of our Dreamer dataset (described in Section~\ref{sec:data}) for this validation.
Instructions are from one of the following classes: Slow down, Speed up, Reach Target Speed, Lane Change, Object centric. 
We evaluate open-loop and use Success Rate as the metric. For each class, we use an individual rule to define success.\looseness=-1

\begin{table}
\footnotesize
\centering
    \setlength{\tabcolsep}{3pt}
    \scalebox{0.85}{\begin{tabular}{l l| l l| c c c } %
        \toprule
         & \textbf{Method} & \textbf{Sensors} & \textbf{Aux. Labels}& \textbf{DS} $\uparrow$ & \textbf{RC} $\uparrow$ & \textbf{IS} $\uparrow$ \\ %
        \midrule
        \parbox[t]{2mm}{\multirow{4}{*}{\rotatebox[origin=c]{90}{Map}}}
        &CaRINA mod.\cite{Rosero2024lrm}	& L,C,M & O & 1.14 &	3.65 &	0.46 	\\ %
        &Kyber-E2E\cite{Elmahgiubi2024kyber} &  L,C,R,M & I, O &	5.47	& 15.89	& 0.34 \\ %
         &TF++\cite{zimmerlin2024tfpp} & L,C & S, D, O, B &5.56&	11.82	& \textbf{0.47} \\
        \midrule
        &\methodBase{}* & C & - & \textbf{ 6.25} & \textbf{18.89} & 0.36 \\
        \midrule
        \midrule
        \parbox[t]{2mm}{\multirow{4}{*}{\rotatebox[origin=c]{90}{Sensor}}}
        &Zero-shot TF++\cite{jaeger2023hidden}	& L,C & S, D, O, B &0.58 &	8.53 &	0.38 \\ %
        & CaRINA hybrid\cite{Rosero2024lrm} & L,C & I, O &	1.23	& 9.56	& 0.31 \\	%
        &TF++\cite{zimmerlin2024tfpp} & L,C & S, D, O, B &5.18	&11.34	&\textbf{0.48}	\\
        \midrule
        &\methodBase{}* & C & - & \textbf{6.87} & \textbf{18.08} & 0.42 \\ %
        \bottomrule
    \end{tabular}}
    \vspace{-0.1cm}
    \caption{\textbf{Leaderboard 2.0 Results.} \methodBase{} achieves state-of-the-art performance on the official Leaderboard 2.0. Legend: L: Lidar, C: Camera, R: Radar, M: Map, priv: privileged, O: Object Detection (3D position and pose), I: Instant Segmentation, S: Semantic Segmentation, D: Depth, B: BEV semantics. *To maintain consistency we changed the naming of our model from CarLLaVA to \methodBase{}.}
    \label{tab:leaderboard}
    \vspace{-0.5cm}
\end{table}

\subsection{Implementation Details}\label{sec:impl}

\boldparagraph{\methodName} Optimization is done with the AdamW optimizer~\cite{Loshchilov2019ICLR} with a weight decay of 0.1 and a learning rate of 3e-5 with a cosine annealing schedule without restarts. We use DeepSpeed v2 for optimizing training efficiency and memory usage. We fully finetune all components besides the LLM for which we use LoRA~\cite{hu2021lora} applied to all linear layers as demonstrated to be effective~\cite{dettmers2024qlora}.
We apply the same data augmentation techniques as TF++~\cite{jaeger2023hidden} but with more aggressive shift and rotation augmentation (shift: 1.5m, rot: 20 deg). 
The model is trained for 14 epochs on 8xA100 80GB GPUs with as batch size of 12 which takes 24 hours. Due to instabilities during training with the L2-Loss after adding additional data we changed to the SmoothL1-Loss.
During inference, we use Commentary by default. This means the final driving model acts in a Chain-of-Thought setting, where first the action and reason are predicted in language space, and then the action is predicted, conditioned on the generated commentary.

\boldparagraph{Baseline - \methodBase{}}
We provide a lightweight baseline driving model that is more suitable for testing driving-specific design choices and evaluating on the official Leaderboard 2.0 in a compute-constrained environment.
A detailed list of the differences to \methodName{} can be found in the Appendix~\ref{sec:supp:comp}, but the main difference is that it has no language support and uses a smaller transformer (50M parameter) based on the LLaMA architecture which is trained from scratch.

\begin{figure}[t]
\begin{center}
   \includegraphics[width=1.0\linewidth]{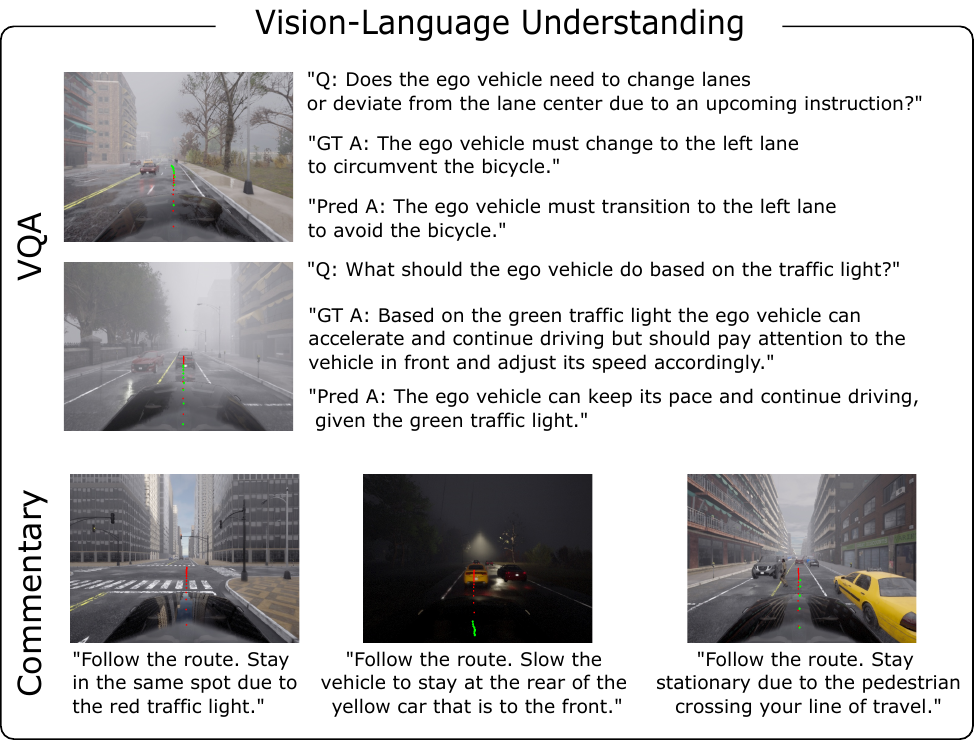}
\end{center}
\vspace{-0.5cm}
\caption{\textbf{Qualitative results for VQA and Commentary.} For VQA we show the question, the ground truth answer and the predicted answer for two examples scenes. Both questions refer to objects far away only apparent in a couple of pixels, but the model still produces correct answers.  }
\label{fig:qual_lang}
\vspace{-0.3cm}
\end{figure}
\begin{figure}[t]
\begin{center}
   \includegraphics[width=1.0\linewidth]{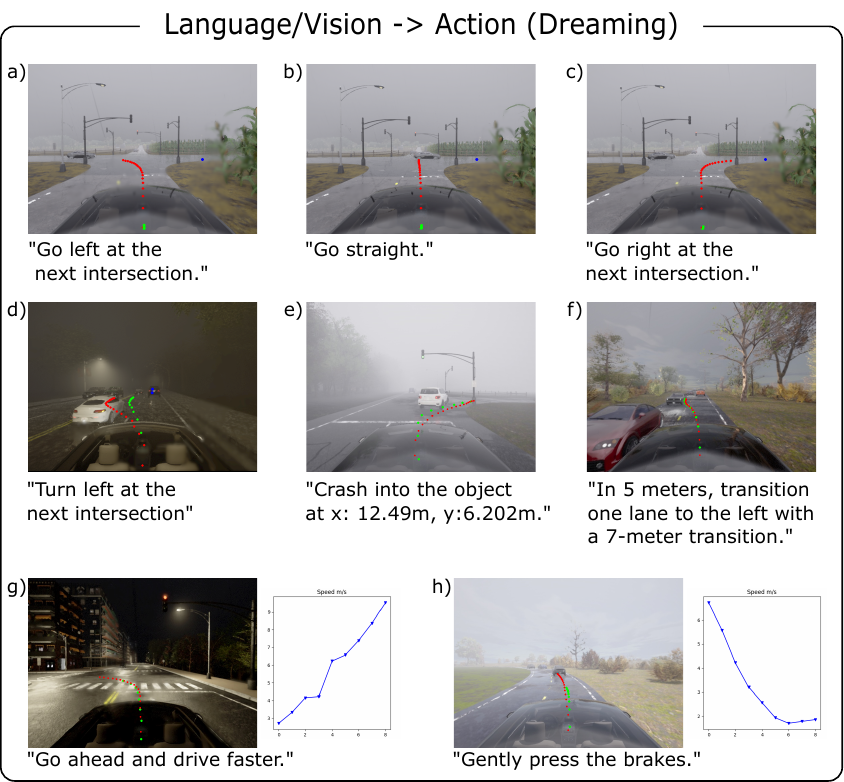}
\end{center}
\vspace{-0.5cm}
\caption{\textbf{Qualitative results of Pose Dreaming.} We show the predicted actions for a diverse set of situations and instructions. The model can successfully adapt to path and speed related instructions. Legend: red: path waypoints, green: speed waypoints, blue graph: speed in m/s.}
\label{fig:qual_dream}
\vspace{-0.5cm}
\end{figure}

\begin{table*}[t!]
\centering
\small
\scalebox{0.8}{\begin{tabular}{l|l|l|c|cccc}
\toprule
& \textbf{Method} & \textbf{Expert} & DS $\uparrow$ & Success Rate(\%) $\uparrow$ & Efficiency $\uparrow$ & Comfortness $\uparrow$ \\ \midrule
\parbox[t]{2mm}{\multirow{5}{*}{\rotatebox[origin=c]{90}{w/ dist.}}} &
TCP~\cite{Wu2022NeurIPS}   &  Think2Drive&  40.70 & 15.00 & 54.26        & 47.80  \\ 
& TCP-ctrl   &  Think2Drive&  30.47 & 7.27 & 55.97        & 51.51 \\
&TCP-traj  &  Think2Drive & 59.90 & 30.00  & 76.54        & 18.08     \\
& ThinkTwice~\cite{Jia2023CVPR}   &  Think2Drive & 62.44 & 31.23 & 69.33 & 16.22   \\
& DriveAdapter~\cite{Jia2023ICCV}  &  Think2Drive & 64.22 & 33.08 & 70.22        & 16.01\\ \midrule \midrule
\parbox[t]{2mm}{\multirow{9}{*}{\rotatebox[origin=c]{90}{w/o dist.}}} &
AD-MLP~\cite{zhai2023ADMLP}   &  Think2Drive &  18.05 & 0.00  & 48.45        & 22.63  \\ 
& UniAD-Tiny~\cite{hu2023_uniad}  &  Think2Drive&  40.73 & 13.18 & 123.92       & 47.04 \\
& UniAD-Base~\cite{hu2023_uniad}  &  Think2Drive & 45.81 & 16.36  & 129.21       & 43.58 \\
& VAD~\cite{jiang2023vad}  &  Think2Drive & 42.35 & 15.00 & 157.94       & 46.01  \\ 
& TCP-traj w/o distillation  &  Think2Drive & 49.30 & 20.45  & 78.78        & 22.96    \\ 
& \new{\qquad our dataset composition}   &  PDM-lite & 45.65 & 18.57  &    74.84    &  \textbf{51.58}  \\ 
& \new{\qquad + with tuned controller}   &  PDM-lite & 63.45 & 37.79  &   228.46     &  30.76  \\ \cmidrule(r){2-7}
& \methodBase{} (LB2.0 model)  & PDM-lite & \textbf{85.94}& 66.82 & 244.18  & 25.49 \\ 
 & \methodName{}  & PDM-lite & 85.07\scriptsize{$\pm$0.95} &	\textbf{67.27\scriptsize{$\pm$2.11}} & \textbf{259.23\scriptsize{$\pm$5.59}} & 33.67\scriptsize{$\pm$5.72}\\
\bottomrule
\end{tabular}}
\vspace{-0.2cm}
\caption{\textbf{Closed-loop Results on Bench2Drive}. Both our models \methodBase{} and the full model \methodName{}, outperforms the previous state of the art by a large margin. We highlight that \methodName{} can preserve the same driving performance as the pure driving model \methodBase{} while including several language related capabilities. \label{tab:bench2drive}}
\vspace{-0.4cm}
\end{table*}

Since the Leaderboard temporarily closed in June 2024 we could not submit the final \methodName{} to the Leaderboard (\methodBase{} was submitted before the closure) and choose to compare the closed-loop performance on a local benchmark (Bench2Drive).
For the Leaderboard model, we apply early stopping during inference to counter the nature of DS described in the metric section. We track the traveled distance and stop driving after a specified distance when the steering angle is close to zero to prevent stopping in the middle of an intersection.\\
\new{\boldparagraph{Baseline - TCP-traj w/o distillation} All existing methods on the Bench2Drive~\cite{jia2024bench} benchmark are trained with a dataset collected by the Think2Drive~\cite{li2024think2drive} expert. Since we use the open-source expert PDM-lite~\cite{Sima2024ECCV} we retrain TCP-traj w/o distillation with our dataset composition collected by PDM-lite to increase fairness in the evaluation. As the driving behavior like maximum speed and acceleration is different between the experts, we also adjust the controller to take those changes into account when evaluating closed-loop. We choose TCP-traj w/o distillation as it is the best previous method that does not rely on expert feature distillation, as expert features are not available for PDM-lite or in the real world.}

\subsection{Results}\label{sec:results}
\boldparagraph{Driving performance on the CARLA Leaderboard 2.0} 
We present the official Leaderboard results in Tab.~\ref{tab:leaderboard}. With our lightweight model \methodBase{} we outperform the previous state of the art (CaRINA hybrid~\cite{Rosero2024lrm}) by 4.6x and the best concurrent work (TF++~\cite{zimmerlin2024tfpp}) by 33\% on the Sensor track. 
\new{\textit{Ablation study:} To show the impact of our output representation, we compare the disentangled path+speed waypoints with the commonly used entangled waypoint representation. We observe a 39.9 percent increase in driving score and a reduction of collisions with static objects to zero. In addition, we analyze the vision encoder and compare it with a ViT without Clip pretraining and with a Resnet-34~\cite{He2016CVPR} pretrained on ImageNet~\cite{Deng2009CVPR} (using the same training budget). The ViT trained from scratch only obtains 0.45 DS, showing the unsurprising capability of the Clip pretraining~\cite{Radford2021ARXIV}. The Resnet results in a driving score of 2.71 compared to our 6.87.}
It is also noteworthy that, to the best of our knowledge, our model is the only model on the leaderboard working only with camera images (note: we did not include entries of the Leaderboard without a method report).
We show more details about the ablations on the output representation, early stopping, and leaderboard variance in the Appendix~\ref{sec:supp:quantitative}.

\boldparagraph{Driving performance on Bench2Drive} 
Tab.~\ref{tab:bench2drive} shows results on the local benchmark Bench2Drive~\cite{jia2024bench} (numbers taken from the newest version of the Bench2Drive Github repository).
Existing works on this benchmark train on data collected with the Think2Drive expert~\cite{li2024think2drive}. However, this expert is not open-source and therefore we cannot collect the data we need to obtain the language annotations. For this reason, we use PDM-lite, an open-source expert for which we can modify the saved labels for our label creation.
\new{To provide a fair comparison and disentangle the performance improvements stemming from the different datasets, we provide results for TCP-traj w/o distillation with our dataset composition and a controller tuned for our dataset. With those two changes, the performance increases from 49.30 DS to 63.45 DS. This shows that the dataset contributes to the performance improvement but our additional advancements are essential to obtain state-of-the-art performance.}
Looking at the driving metrics (DS and SR) \methodName{} and \methodBase{}, which is the identical model that we used for the Leaderboard, outperforms all previous methods by a large margin. For all \methodName{} models, we train three seeds to average out the training variance. 
Only in the comfortness metric previous works partly perform better\new{, which is likely due to the higher driving speed of our method.}
The Vision-Language-Action model (\methodName{}) can preserve the driving performance of the pure driving model (\methodBase{}) while including a wide range of language abilities.

\begin{table}[t!]
\centering
\small
\resizebox{0.4\textwidth}{!}{
\begin{tabular}{l|cccc}
\toprule
\multirow{2}{*}{Model}   & \multicolumn{2}{c}{DriveLM-VQA}  & \multicolumn{2}{c}{Commentary} \\ \cmidrule(r){2-3} \cmidrule(r){4-5}
                                &  GPT $\uparrow$  & SPICE $\uparrow$ &  GPT $\uparrow$  & SPICE $\uparrow$  \\ \midrule
        InternVL2-1B     & 33.08 & 30.55 & 14.95 & 7.60\\ 
        InternVL2-2B    & 31.22 & 44.40 & 20.53 & 9.04\\ 
        InternVL2-4B    & 27.11 & 43.51 & 24.75 & 8.12 \\ 
        \methodName{}-1B   & 58.48 & \textbf{56.77} & \textbf{78.94} & \textbf{38.04} \\
        \bottomrule
\end{tabular}
}
\vspace{-0.2cm}
\caption{\textbf{Language ability.} We show results on DriveLM and our Commentary benchmark on a balanced evaluation split. \methodName{} outperforms the InternVL2 models which we evaluate zero-shot.
}
\vspace{-0.5cm}
\label{table:vqa}
\end{table}

\boldparagraph{Language understanding - VQA \& Commentary} 
We show results on the DriveLM-hard and Commentary Benchmark in Tab.~\ref{table:vqa}. We use the Mini-InternVL2 models to provide a baseline comparison. However, we emphasize that \methodName{} was trained on the simulated visual in-distribution data, whereas InternVL2 is tested zero-shot. The baseline models get basic questions right like identifying a traffic light or a stop sign but struggle with more complex, driving-specific questions. \new{In addition, we compare to a version that is fine-tuned only on DriveLM (InternVL2-1B-DriveLM). This specialized model obtains a GPT-score of 63.85, which is only slightly better than our model that is able to perform several different tasks.}
We provide qualitative examples in Fig.~\ref{fig:qual_lang} and the Appendix~\ref{sec:supp:qualitative}. We found that the model especially struggles with answer types that are rarely seen during training indicating that either collecting more driving samples with those answer types or oversampling during training might help to further improve the performance.

\begin{table}[t!]
\centering
\small
\resizebox{0.3\textwidth}{!}{
\begin{tabular}{l|cc}
\toprule
                                &  DS $\uparrow$ & SR(\%) $\uparrow$  \\
        \toprule
        GPS target point (TP)  & 85.07\scriptsize{$\pm$0.95} &	\textbf{67.27\scriptsize{$\pm$2.11}} \\
        Language command (HLC) &  \textbf{86.08\scriptsize{$\pm$1.76}} &	65.78\scriptsize{$\pm$3.90} \\
        \bottomrule
\end{tabular}
}
\vspace{-0.2cm}
\caption{\textbf{Following navigational commands.} \methodName{} obtains strong driving results independent of the navigational conditioning (GPS target points vs. language command) showing it's ability to follow instructions.}
\label{table:hlc}
\vspace{-0.2cm}
\end{table}
\boldparagraph{Following navigational commands}
To show the ability of the model to follow basic navigational commands we compare \methodName{} with navigational conditioning of GPS target points with the same model using language commands (HLC) (i.e., language instructions like \textit{turn right}). Tab.~\ref{table:hlc} shows that by only using HLC we reach a driving performance, which is within the variance of the results with using TPs.
This enables more human-like conditioning but also indicates that the shortcut of recovery through target point locations~\cite{jaeger2023hidden} is not needed in our setting.

\begin{table}[t!]
\centering
\footnotesize
\resizebox{0.45\textwidth}{!}{
\begin{tabular}{l|ccccc|c}
\toprule
                   Metric         & \multicolumn{6}{c}{ SR(\%) $\uparrow$}  \\ 
                   Category         & Faster & Slower & Target Speed & Lane Change & Objects & Avg. \\
                            \cmidrule{1-1} \cmidrule{2-6} \cmidrule{7-7} 
        w/o Dreamdata     &  56.45 & 22.58 & 19.35 & 3.23 & 20.97 & 24.52 \\
        \methodName{}       & 92.45 & 84.91 & 86.79 & 83.02 & 58.49 & 81.13 \\
        \bottomrule
\end{tabular}
}
\vspace{-0.2cm}
\caption{\textbf{\dream{}.} When activating the dreaming flag, \methodName{} is able to follow a wide range of instructions compared to the model trained with only the natively supported commands from the CARLA simulator.}
\label{table:dream}
\vspace{-0.5cm}
\end{table}

\boldparagraph{\dream{}}
For the general ability to align the language understanding with the action space, we evaluate on the \dream{} dataset.
Tab.~\ref{table:dream} shows the Success Rate of the \dream{} evaluation for each category. We compare the model trained on the \textit{Dreaming} data with a baseline that is only trained on instructions provided by the CARLA simulator. With training on the synthetically generated instruction-action pairs, we see a huge improvement (28.22 to 72.96 SR).
Fig.~\ref{fig:qual_dream} shows qualitative examples of the Pose Dreaming mode. In the first row (a-c), we show navigational changes for the same scene. This shows that the model does not overfit to visual cues but instead pays attention to the language. In a) and d) we use the same instruction of turning left but in a different context indicating that the model can successfully adapt its response to the situation. In scene a) we are right before a junction so the model correctly predicts a left turn indicated by the red path waypoints. In contrast, in scene d) we are on a multi-lane road far from a junction so \textit{turning left} means to prepare for the turn and move into the left lane to perform the turn when a junction shows up. More examples, including failure cases, can be found in the Appendix~\ref{sec:supp:qualitative}.

\begin{table}[tb!]
\centering
\small
\scalebox{0.8}{\begin{tabular}{l|ccc|cc}
\toprule
  &  \multirow{2}{*}{\rotatebox{90}{\textbf{Comm}}}& \multirow{2}{*}{\rotatebox{90}{\textbf{VQA}}} &  \multirow{2}{*}{\rotatebox{90}{\textbf{Dream}}}& \multicolumn{2}{c}{\textbf{Bench2Drive}} \\ \cmidrule{5-6} 
 & & & & DS $\uparrow$ & SR(\%) $\uparrow$ \\ \midrule
\parbox[t]{2mm}{\multirow{4}{*}{\rotatebox[origin=c]{90}{\methodName{}}}}
 & \checkmark &\checkmark & \checkmark & 85.07\scriptsize{$\pm$0.95} &	67.27\scriptsize{$\pm$2.11} \\
 & \checkmark & \checkmark &- & 84.41\scriptsize{$\pm$1.38} & 64.85\scriptsize{$\pm$2.24}\\
  & \checkmark & - &- & 84.55\scriptsize{$\pm$0.42} & 65.00\scriptsize{$\pm$0.64}\\
 & - & - &- & 84.67\scriptsize{$\pm$1.52} & 64.70\scriptsize{$\pm$3.22}\\
\bottomrule
\end{tabular}}
\vspace{-0.2cm}
\caption{\textbf{Closed-loop results on Bench2Drive for different training mixtures}. Pure vision-language tasks (Commentary and VQA) do not influence pure driving performance (last row). But including \dream{} training slightly improves driving. \label{tab:b2dabl}}
\vspace{-0.5cm}
\end{table}

\boldparagraph{Training mixture}
We ablate every added language task from the full \methodName{} model to a base driving model without any language capabilities. Tab~\ref{tab:b2dabl} shows that having a pure driving model (4th row) gets a similar performance as the model with (non-aligned) language understanding (2nd + 3rd rows). Hence, in our experiments, we have not seen any positive or negative impact from not-aligned language tasks on driving. However, adding the \dream{} data to the training mixture slightly improves driving performance on the Bench2Drive benchmark.

\section{Conclusion and Limitations}
We present \methodName{}: a VLM-based autonomous driving model that achieves state-of-the-art results on the official CARLA Leaderboard 2.0 and Bench2Drive while demonstrating language understanding. We measure language performance on driving Commentary and VQA showing that after finetuning, a base generalist VLM (InternVL2) can excel in a specialized driving domain. To align this vision-language understanding with the action space we introduce the task of \dream{}, where we demonstrate a high success rate of the model to predict actions for a high variety of language instructions.

\boldparagraph{Limitations}
We recognize several limitations of our work. For technical and organizational reasons (Leaderboard closed in June 2024), we were able to test only \methodBase{} on the official CARLA Leaderboard 2.0 and not the full \methodName{} model, which we leave for future work. However, we rigorously checked the closed loop driving performance of \methodName{} and \methodBase{} on the local Bench2Drive benchmark. 
Also, while we use Chain-of-Thought (CoT) (i.e., conditioning the driving action on the intermediate commentary), we have not yet observed statistically significant driving improvements (Appendix~\ref{sec:supp:cot}). We hypothesize that appropriate CoT-specific data and training recipes are needed to harness its benefits.
Finally, we perform driving and language understanding only in simulation. We appreciate that adding VLMs in real-world driving models increases the inference latency, but we believe that advances in smaller VLMs and engineering efforts (such as \cite{lingo2, nuro}) would allow testing VLMs real-time on the car.

\vspace{0.2cm}
\new{\noindent \textbf{Acknowledgements.}
We thank the whole Wayve Lingo team, Ana-Maria Marcu, Jan Hünermann, Benoit Hanotte, Alice Karnsund and Jamie Shotton for helpful discussions and proofreading.
We also thank Kashyap Chitta, Julian
Zimmerlin, Jens Beißwenger, Bernhard Jäger and Andreas
Geiger for valuable discussions and help with the expert. We thank the International Max Planck Research
School for Intelligent Systems (IMPRS-IS) for supporting
K. Renz.}
{
    \small
    \bibliographystyle{ieeenat_fullname}
    \bibliography{bibliography_short, bibliography_custom, bibliography}
}

\clearpage
\setcounter{page}{1}
\renewcommand{\thesubsection}{\Alph{subsection}}
\setlist[itemize]{leftmargin=1cm, itemsep=0pt, topsep=0pt, parsep=0pt, partopsep=0pt}

\subsection{Datasets}\label{sec:supp:datasets}
We provide a more detailed description of the collected datasets and how we generate the labels for each language-related dataset.
\subsubsection{Driving dataset - Scenarios}
In each Town, we collect data containing different scenarios, which we detail in the following (descriptions are taken from \url{https://leaderboard.carla.org/scenarios/}):
\begin{itemize}
    \item \textbf{Control Loss without Previous Action:}  
    The ego-vehicle loses control due to poor road conditions and must recover.

    \item \textbf{Unprotected Left Turn at Intersection with Oncoming Traffic:}  
    The ego-vehicle performs an unprotected left turn at an intersection (can occur at both signalized and unsignalized intersections).

    \item \textbf{Right Turn at Intersection with Crossing Traffic:}  
    The ego-vehicle makes a right turn at an intersection while yielding to crossing traffic (signalized and unsignalized intersections).

    \item \textbf{Crossing Negotiation at Unsignalized Intersection:}  
    The ego-vehicle navigates an unsignalized intersection by negotiating with other vehicles. The assumption is that the vehicle entering the intersection first has priority.

    \item \textbf{Crossing Traffic Running a Red Light at an Intersection:}  
    While the ego-vehicle is traveling straight through an intersection, a crossing vehicle runs a red light (signalized and unsignalized intersections).

    \item \textbf{Crossing with Oncoming Bicycles:}  
    The ego-vehicle must turn at an intersection while yielding to bicycles crossing the intersection.

    \item \textbf{Highway Merge from On-Ramp:}  
    The ego-vehicle merges into moving traffic on a highway.

    \item \textbf{Highway Cut-In from On-Ramp:}  
    A vehicle merges into the ego-vehicle’s lane from an on-ramp. The ego-vehicle must decelerate, brake, or change lanes to avoid a collision.

    \item \textbf{Static Cut-In:}  
    Another vehicle cuts into the ego lane from a queue of stationary traffic. It must decelerate, brake, or change lanes to avoid a collision.

    \item \textbf{Highway Exit:}  
    To exit the highway the ego-vehicle needs to cross a lane of moving traffic.

    \item \textbf{Yield to Emergency Vehicle:}  
    An emergency vehicle is approaching from behind. The ego must create space for it to pass safely.

    \item \textbf{Obstacle in Lane - Same Direction:}  
    An obstacle (e.g., a construction zone, an accident, or a parked vehicle) is blocking the ego lane. The ego vehicle must change lanes into traffic moving in the same direction to bypass the obstacle.

    \item \textbf{Obstacle in Lane - Opposite Direction:}  An obstacle (e.g., construction zone, an accident, or a parked vehicle) is blocking the ego lane. The ego vehicle must change lanes into traffic moving in the opposite direction to bypass the obstacle.

    \item \textbf{Door obstacle:} The ego-vehicle needs to avoid a parked vehicle with its door opening into the lane.

    \item \textbf{Slow moving hazard at lane edge:} A slow-moving hazard (e.g. bicycle) partially obstructs the ego vehicle’s lane. The ego-vehicle must either brake or carefully bypass the hazard (bypassing on lane with traffic in the same or opposite direction).

    \item \textbf{Vehicle invading lane on bend:} On a bend, an oncoming vehicle invades the ego vehicle’s lane to avoid an obstacle. The ego-vehicle must brake or move to the side of the road to safely navigate past the oncoming vehicle.

    \item \textbf{Longitudinal control after leading vehicle’s brake:} The leading vehicle in front of the ego-vehicle brakes suddenly to avoid an obstacle. The ego-vehicle must execute an emergency brake or avoidance maneuver to prevent a collision.

    \item \textbf{Obstacle avoidance without prior action:} The ego-vehicle encounters an unexpected obstacle or entity on the road. It must perform an emergency brake or avoidance maneuver.

    \item \textbf{Pedestrian emerging from behind parked vehicle:} A pedestrian suddenly emerges from behind a parked vehicle and enters the lane. The ego-vehicle must brake or take evasive action to avoid hitting the pedestrian.

    \item \textbf{Obstacle avoidance with prior action - pedestrian or bicycle:} While in the middle of a turn, the ego-vehicle encounters an obstacle such as a pedestrian or a bicycle crossing the road or a stopped vehicle in the road and must perform an emergency brake or avoidance maneuver.

    \item \textbf{Parking Cut-in:} A parked vehicle exits a parallel parking space into the ego-vehicle’s path. The ego-vehicle must slow down to allow the parked vehicle to merge into traffic.

    \item \textbf{Parking Exit:} The ego-vehicle must exit a parallel parking space and merge into moving traffic.
\end{itemize}

\subsubsection{VQA - DriveLM}
For the VQA data, we use the DriveLM-Carla~\cite{Sima2024ECCV} data generation method. Since we generate a new driving dataset, we extract questions and answers for our dataset instead of using the original dataset. For the training set, we generate in total 28M QA-pairs for 1M frames of Town 12. For the evaluation set, we use the keyframe extraction of DriveLM to evaluate on more interesting and less redundant frames. In addition, we balance the validation set to capture the same amount of samples for each answer type in the dataset.\\
Since the labels are auto-generated with a heuristic-based procedure, the QAs follow the same sentence structures and wordings. To avoid overfitting to specific phrases we include data augmentation. For this, we prompt GPT-4 to generate 20 alternative sentences for each question and answer, from which we sample during data loading.

\subsubsection{Commentary}
We generate language labels for the \textit{Commentary} task based on a subset of the simulator state, which we save during the data collection. 
The structure of the \textit{Commentary} labels is as follows: The first sentence describes the action according to the route with its justification (e.g., staying on the current lane, doing a lane change to go around an obstacle). It is followed by a description of the speed-related action (e.g., accelerate, keep the speed, stop), and the reason (e.g., because of the pedestrian, to follow the vehicle in front, to drive closer to the stop sign). \\
In the following, we detail the steps to obtain each part of the \textit{Commentary} labels. \\
\boldparagraph{Route action}
The default description is ``\textit{Follow the route.}". Only in special cases, we change the description. For this, we check if any scenario is active in the given frame and get the scenario type. We only change the default description for the scenarios requiring a deviation from the center lane of the original route (e.g., obstacle in lane, vehicle invading the lane, door obstacle). Since the ego action differs depending on the location relative to the scenario, we extract the relative positioning from the simulator information. Those locations are grouped into three phases: (1) before the lane deviation, (2) during the deviation, and (3) end of the deviation. 
\textit{Before the deviation}, we differentiate between the scenario types and use a template sentence for each, for instance:
\begin{itemize}
    \item Overtake the bikes on your lane.
    \item Go around the vehicle with the open door.
    \item Give way to the emergency vehicle.
    \item Go around the accident in your lane.
    \item Go around the construction site.
    \item Move slightly to the right to circumvent the oncoming cars entering your lane because of the construction cones.
\end{itemize} 
\textit{During the deviation}, describes the phase in which the ego already shifted lanes. We reuse the templates from (1) but add ``Stay on your current lane to" before the templates (e.g., Stay on your current lane to overtake the bikes on your lane.) \\
\textit{End of the deviation} is the phase where the ego vehicle needs to shift back to its original lane. Since our model is based on only front-view cameras we use a generic sentence for this (i.e., ``Return to your original route after avoiding the obstacle.") as often the type of the obstacle or scenario is not visible anymore. This template can be easily changed for a model supporting multi-view inputs.

\boldparagraph{Speed action}
We generate a high-level description of the ego action based on the current speed, the desired target speed based on the expert decision, and the current speed limit. We differentiate between the following types:
\begin{itemize}
    \item Remain stopped
    \item Come to a stop now
    \item Maintain your current speed
    \item Maintain the reduced speed
    \item Increase your speed
    \item Slow down
\end{itemize} 
For the scenarios that are used for the \textit{route action} description (i.e., where the expert needs to deviate from the route), we use a different sentence template. This is only the case for the situation when the ego vehicle is \textit{before the deviation} and is stopped and remains stopped for the next two seconds. In this case, we use the template ``Wait for a gap in the traffic before changing lanes".

\boldparagraph{Speed Reason}
Next, we leverage the Inteligent-Driver-Model (IDM) features, which the expert algorithm is based on. IDM identifies leading objects and calculates the optimal target speed for the ego vehicle based on the distance to the leading object. Leading objects include dynamic objects like other vehicles or pedestrians and static objects like traffic lights, stop signs, or construction sites.
With this, we know for any sample in the dataset which object the main reason is for the given target speed and therefore the \textit{speed action}. Based on the type of the leading object, we construct a language description of the object. For vehicles, this consists of the color of the vehicle, the type (e.g., SUV, police car), and a rough position relative to the ego (e.g., to the front right). For static objects, it is the name of the object (e.g. traffic light) and in case the object has a state this is also included (e.g., \textit{red} traffic light). For pedestrians, we differentiate between children and adults.
Based on the type of the leading object and the \textit{speed action} we construct different sentences, for instance:
\begin{itemize}
    \item since you've already stopped at the stop sign
    \item to avoid a collision with the \textit{object description}
    \item due to the pedestrian crossing in front of you
    \item to remain behind the red SUV that is slowing down because of a red traffic light.
    \item to reach the target speed
    \item because the traffic light is green
\end{itemize} 

If we are right before a junction we add another notice label regarding the positioning of the other vehicles in the junction. With this, the model needs to reason about whether it is safe to enter a junction or not. We collect the position and driving direction of each vehicle close to the junction and summarize the situation based on one of the following sentences:
\begin{itemize}
    \item the other vehicles are stopped at the junction and the junction is clear
    \item the other vehicles are stopped at the junction and the vehicle in the junction is moving away
    \item pay attention to the vehicles coming towards the junction
    \item pay attention to the vehicle in the junction
\end{itemize}

\subsubsection{Action Dreamer}

We construct an offline, non-reactive simulation based on the collected dataset to generate alternative ego trajectories and evaluate their feasibility in terms of collision avoidance and adherence to traffic rules. For this purpose, we utilize the Kinematic Bicycle Model in combination with the PID controllers from the PDM-lite expert algorithm.\\
The core functionality of the Action Dreamer simulation is the \textit{ego forecasting}, which predicts future ego vehicle poses based on the ego actions in each timestep. There are several approaches to generate these ego actions, allowing for modification to obtain the alternative trajectories. One approach involves perturbing the ground truth actions to produce slightly modified trajectories. Another approach uses the PID controllers to compute actions based on pre-defined path waypoints and target speeds. In this case, the lateral PID controller generates steering angles, while the longitudinal PID determines acceleration and braking values based on the desired target speed. Using these actions, the Kinematic Bicycle Model calculates the next vehicle pose. This process can be iteratively unrolled over multiple time steps to derive a complete trajectory. \\
We start with obtaining the current state of the simulator from the saved dataset. For each dynamic object, we also get the states for the following 10 timesteps. With this we can get the non-reactive trajectories for each object and perform collision checks with the ego vehicle.
As default, we use the ground truth actions of the ego vehicle.
In addition, we use the ground truth path provided by the experts' path planner, which includes waypoints spaced every 10 cm, as the default route to be followed. We then change those default values to obtain alternative trajectories for the modes: objects (collision), faster, slower, target speed, and lane changes. \\
The following steps describe how we obtain the necessary information (e.g., actions, updated paths with desired speeds, or a combination of both) for the ego forecasting method for each of the different modes we have.

\begin{itemize}
    \item \textbf{Objects (Collision) Mode:}  
    Filter all dynamic and static objects, retaining only those within a 15-meter radius of the ego route and at least 3 meters ahead of the ego vehicle. For each object, calculate its position, distance to the ego vehicle, and whether the ego vehicle (given its current speed) could reach the object within a given future timestep. Objects that are too far and unreachable are discarded. For reachable objects, we calculate the target speed required for the ego vehicle to precisely reach the object’s location in the required time. For the path, we adjust the route waypoints to ensure they intersect with the target object’s position. This modified route, along with the computed target speed, is then passed to the ego forecasting process.
    
    \item \textbf{Faster Mode:}  
    We use the original path and actions for steering but set the acceleration to a random value above 50\%.
    
    \item \textbf{Slower Mode:}  
    Similar to the faster mode but we set the acceleration to zero and activate the brake.
    
    \item \textbf{Random Target Speed:}  
    We assign a random target speed between 0 and 35 m/s and directly pass this together with the default path to the forecasting method.
    
    \item \textbf{Lane Changes:}  
    We exclude frames where the vehicle is already performing a lane change or is in a junction. To obtain the number of possible options, we extract information on the number and types of lanes (e.g., driving lanes, parking lanes, or sidewalks) in both the same and opposite directions. For each of the options, we change the default path so that it reaches the specified lane. We randomize the starting distance of the lane change and the length of the transition phase. Those parameters are conditioned on the current ego speed. 
\end{itemize}

\begin{figure*}[t]
\begin{center}
   \includegraphics[width=0.575\linewidth]{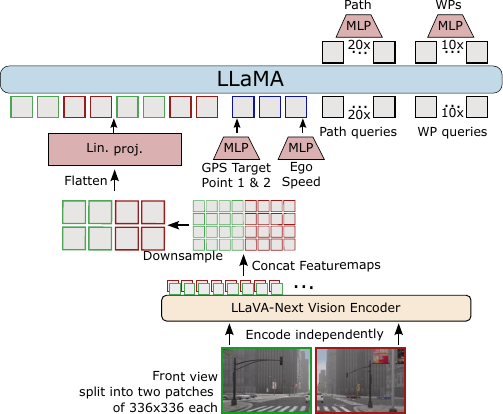}
\end{center}
\vspace{-0.4cm}
\caption{\textbf{\methodBase{} architecture.} The images are split in two, and each split is independently encoded and then concatenated, downsampled, and projected before feeding it into a transformer decoder which is based on the LLaMA architecture. The output utilizes the same output representation as \methodName. }
\vspace{-0.2cm}

\label{fig:architecture2}
\end{figure*}

\subsection{Implementation details}\label{sec:supp:implementation}
\subsubsection{Hyperparameter} 
Table~\ref{tab:hyper} shows the hyperparameter we use to train \methodName{}.

\begin{table}[tb!]
\centering
\small
\scalebox{1.0}{\begin{tabular}{l|l}
\toprule
Epochs & 14 \\
Learning Rate & 3e-5\\
Batch Size & 96 \\
Optimizer & AdamW\\
Weight decay & 0.1\\
Betas & (0.9, 0.999)\\
LR schduler & One cycle cosine\\
Warmup steps & 5\% of total steps\\
LoRA alpha & 64 \\
LoRA r & 32 \\
LoRA dropout & 0.1 \\
\bottomrule
\end{tabular}}
\caption{\textbf{Hyperparameter} choices to train \methodName{}.  \label{tab:hyper}}
\vspace{-0.4cm}
\end{table}

\subsubsection{Training buckets}\label{sec:supp:buckets}
The majority of driving involves straight, uneventful segments. To maximize the collection of interesting scenarios during data collection, we focus on capturing a diverse range of challenging situations. However, some ratio of easy and uneventful data is inevitable. Training models on the entire dataset revealed that straight driving without hazards is effectively learned in the early epochs, resulting in wasted computation in later epochs as the models continue to train on these uninteresting samples. To address this issue, we create data buckets containing only the interesting samples and sample from these buckets during training instead of the entire dataset. 
We use (1) five buckets for different amounts of acceleration and deceleration with one specifically for starting from stop, excluding samples with an acceleration between -1 and 1, (2) two buckets for steering, excluding samples for going straight, (3) three buckets for vehicle hazard with vehicles coming from different directions, (4) one for a stop sign, red light, and walker hazards each, (5) one bucket for swerving around obstacles, (6) one bucket for ``old" Towns commonly used in Leaderboard 1.0 models (Town 01-10) and (7) one bucket that samples from the whole dataset to keep a small portion of uneventful data such as driving straight.

\subsubsection{\methodName{} vs. \methodBase{}}\label{sec:supp:comp}

Our base model \methodBase{} was designed as a lightweight model to research driving-specific design choices. With adding language capabilities, we also changed some of the settings to better match the added requirements.  We note that because of the closure of the CARLA Leaderboard and because of computational overhead we did not repeat the experiments of the \methodBase{} with the new settings. We believe that the claims and results can be expected to still hold. 
Fig.~\ref{fig:architecture2} shows the architecture of \methodBase{}. We detail the exact differences between \methodBase{} and \methodName{}:
\begin{enumerate}
    \item Number of epochs: \methodBase{} is trained for 30 epochs. For \methodName{} we reduce the number of epochs to 14. 
    \item Image encoder: \methodBase{} uses a Clip-ViT that is used as default in the LLaVA VLM. \methodName{} uses the original image encoder of the InternVL2-1B. In both cases, we do full finetuning.
    \item Language model: \methodBase{} uses a 50M parameter transformer decoder based on the LLaMA architecture which we train from scratch. \methodName{} uses the default pretrained LLM from the InternVL2-1B model which we finetune with LoRA.
    \item Loss function: \methodBase{} uses L2-Loss. After adding the additional \dream{} data we observed instabilities during training with the L2-Loss so we changed to the SmoothL1-Loss.
    
\end{enumerate}

\subsubsection{Metric descriptions}\label{sec:supp:metric}
\boldparagraph{Leaderboard 2.0} \\
The Leaderboard uses the official CARLA metrics: Driving Score, Route Completion, and Infraction Score. Each metric is calculated for each route independently. After all routes are completed, the final metrics are derived by taking the arithmetic mean of the metrics across all routes. The overall driving score, calculated using the global values, is the primary metric for ranking methods.\\
\textit{Driving Score.} 
The primary evaluation criterion is the \textit{Driving Score}, denoted as:
\[
DS_i = RC_i \cdot IS_i,
\]
where \( RC_i \) represents the percentage of the \( i\)-th route completed, and \( IS_i \) is a penalty factor accounting for infractions incurred during the route. \\
\textit{Route Completion.}
This metric quantifies the proportion of the route successfully completed by the agent, expressed as a percentage.\\
\textit{Infraction Penalty.}
The penalty due to infractions, \( IS_i \), is calculated as a product of all infractions:
\[
IS_i = \prod_{j=1}^{N_{I}} (p_j)^{\#\text{infractions}_j},
\]
where \( p_j \) denotes the penalty coefficient for the \( j\)-th type of infraction out of a total of \( N_{I} \) infraction types, which we specify int following. \( \#\text{infractions}_j \) is the number of times this infraction occurred. The calculation begins with a base score of 1.0, which decreases with each infraction. \\
Infractions are categorized by severity, each associated with a penalty coefficient that reduces the driving score. Key infractions include:
\begin{itemize}
    \item \textbf{Collisions with pedestrians:}  \( p_j = 0.50 \).
    \item \textbf{Collisions with vehicles:}  \( p_j = 0.60 \).
    \item \textbf{Collisions with static objects:}  \( p_j = 0.65 \).
    \item \textbf{Running a red light:}  \( p_j = 0.70 \).
    \item \textbf{Ignoring a stop sign:}  \( p_j = 0.80 \).
    \item \textbf{Failure to yield to emergency vehicles:} \( p_j = 0.7 \).
    \item \textbf{Failure to maintain minimum speed:} Up to \( p_j = 0.7 \).
    \item \textbf{Off-road driving:} Reduces route completion score proportionally.
\end{itemize}
When one of the following events occurs, the route stops immediately:
\begin{itemize}
    \item Route deviation (more than 30 meters off route).
    \item Blocked agent (more than 180 simulation seconds without action).
    \item Communication timeout (more than 60 seconds).
    \item Route timeout (exceeding allowed simulation time).
\end{itemize}

\boldparagraph{Leaderboard 2.0 metric discussion} \\
The Driving Score is calculated in a way, that it can be advantageous not completing the whole route.
This is the case if the infractions incurred during a segment of the route reduce the driving score more than the potential gain from continuing the route. In this case stopping early to avoid further penalties leads to an overall higher driving score. This tradeoff only occurs for long routes.
We refer to~\cite{Zimmerlin2024TECH} for a mathematical description of this tradeoff, a detailed discussion and a proposal for a better metric calculation.

\boldparagraph{Bench2Drive} \\
\textit{Driving Score.}
The Driving Score is calculated similarly to the Leaderboard 2.0. The only difference to the original Driving Score is that the penalty ``Failure to maintain minimum speed" is ignored. To take the driving speed into account the authors of the benchmark introduced the ``Efficiency" metric. Since Bench2Drive uses short routes the discussed trade-off does not occur on this benchmark.\\
\textit{Success Rate.}
The Success Rate measures the percentage of completed routed without any infractions (ignoring the minimum speed penalty).\\
\textit{Efficiency.} This uses the ratio of the ego vehicle speed to the speed of the surrounding actors. Since there was no penalty in Leaderboard 1.0 for low speeds most models used a very low speed, which makes driving and reacting to other dynamic actors much easier. The higher this efficiency metric the faster the model drives, making the driving task harder. \\
\textit{Comfortness.} The comfortness metric takes the jerk magnitude, lateral and longitudinal acceleration, yaw acceleration, longitudinal jerk, and the yaw rate into account. If the mean of the ego vehicles measurement over the full route falls into the following thresholds it is treated as success. 
\begin{itemize}
    \item \textbf{Jerk Magnitude:} Maximum absolute magnitude of jerk is \( 8.37 \, \text{m/s}^3 \).
    \item \textbf{Lateral Acceleration:} Maximum absolute lateral acceleration is \( 4.89 \, \text{m/s}^2 \).
    \item \textbf{Longitudinal Acceleration:} 
    \begin{itemize}
        \item Maximum longitudinal acceleration is \( 2.40 \, \text{m/s}^2 \).
        \item Minimum longitudinal acceleration is \( -4.05 \, \text{m/s}^2 \).
    \end{itemize}
    \item \textbf{Yaw Acceleration:} Maximum absolute yaw acceleration is \( 1.93 \, \text{rad/s}^2 \).
    \item \textbf{Longitudinal Jerk:} Maximum absolute longitudinal jerk is \( 4.13 \, \text{m/s}^3 \).
    \item \textbf{Yaw Rate:} Maximum absolute yaw rate is \( 0.95 \, \text{rad/s} \).
\end{itemize}
The thresholds are taken from the Bench2Drive repository.

\begin{table*}[t!]
\centering
\scalebox{0.9}{\begin{tabular}{l|l|ccccc|c}
\toprule
& \multirow{2}{*}{\textbf{Method}} & \multicolumn{5}{c}{\textbf{Ability} (\%) $\uparrow$}                                                                                                                \\ \cmidrule{3-8} 
                                 & & \multicolumn{1}{c}{Merging} & \multicolumn{1}{c}{Overtaking} & \multicolumn{1}{c}{Emergency Brake} & \multicolumn{1}{c}{Give Way} & Traffic Sign & \textbf{Mean} \\ \midrule
\parbox[t]{2mm}{\multirow{5}{*}{\rotatebox[origin=c]{90}{w/ dist.}}} & TCP~\cite{Wu2022NeurIPS}                & 16.18           & 20.00               & 20.00                    & 10.00             & 6.99                  & 14.63         \\
& TCP-ctrl          & 10.29             & 4.44                & 10.00                    & 10.00             & 6.45                  & 8.23        \\
& TCP-traj          & 8.89            & 24.29               & 51.67                    & 40.00             & 46.28                & 28.51         \\
& ThinkTwice~\cite{Jia2023CVPR}        & 27.38           &         18.42      & 35.82                    & 50.00             & 54.23                 & 37.17       \\
& DriveAdapter~\cite{Jia2023ICCV}       & 28.82           & 26.38             & 48.76                    & 50.00             & 56.43                 & 42.08         \\ \midrule
\parbox[t]{2mm}{\multirow{7}{*}{\rotatebox[origin=c]{90}{w/o dist.}}} & AD-MLP~\cite{zhai2023ADMLP}             & 0.00             & 0.00                & 0.00                     & 0.00              & 4.35                 & 0.87         \\
& UniAD-Tiny~\cite{hu2023_uniad}          & 8.89             & 9.33              & 20.00                    & 20.00             & 15.43                & 14.73        \\
& UniAD-Base~\cite{hu2023_uniad}          & 14.10           & 17.78               & 21.67                   & 10.00             & 14.21               & 15.55         \\
& VAD~\cite{jiang2023vad}                & 8.11             & 24.44               & 18.64                    & 20.00             & 19.15                 & 18.07        \\ 
& TCP-traj w/o distillation & 17.14   & 6.67             & 40.00                  & 50.00             & 28.72                & 28.51         \\ \cmidrule(r){2-8}
& \methodBase{} (LB2.0 model) & \textbf{60.00} & \textbf{60.00} & 78.33 & 50.00 & 77.89 & 65.25 \\
& \methodName{} & 54.01\scriptsize{$\pm$2.63} &	57.04\scriptsize{$\pm$3.40} &	\textbf{88.33\scriptsize{$\pm$3.34} }&	\textbf{53.33\scriptsize{$\pm$5.77}} &	\textbf{82.45\scriptsize{$\pm$4.73}} &	\textbf{67.03\scriptsize{$\pm$2.12}} \\ 
\bottomrule
\end{tabular}}
\vspace{-0.2cm}
\caption{\textbf{Multi-Ability Results of Bench2Drive.} We outperform the existing methods in all abilities and can improve in average by 25 percentage points.  \label{tab:ability}}
\vspace{-0.2cm}
\end{table*}

\boldparagraph{Vision-Language Understanding} \\
For the tasks of VQA and Commentary, we use SPICE~\cite{anderson2016spice} and DriveLM's GPT Score~\cite{Sima2024ECCV} as metrics.
SPICE is a metric used for image captioning with a higher correlation to human judgment than other automatic language metrics like Cider~\cite{vedantam2015cider} or Meteor~\cite{lavie-agarwal-2007-meteor}.
The GPT Score is based on the DriveLM implementation with two smaller changes. Since we could not directly compare to their numbers, because of a different evaluation set those changes should not have any impact on the conclusions drawn. 
The first change is using GPT-4 (gpt-4o-2024-08-06) instead of GPT-3.5. In addition, we add ``Just rate the similarity of the content not the sentence structure. If the content is completely different rate with 0." to the prompt as we found this to be more accurate.

\begin{table}[t]
\vspace{-0.5cm}
\centering
\footnotesize
\subfloat[
\textbf{Output.}
\label{tab:output}
]{
\centering
\begin{minipage}{0.38\linewidth}{\begin{center}
\begin{tabular}{l|r r}
\toprule
& \multicolumn{1}{c}{\textbf{DS} $\uparrow$} & \textbf{Stat} $\downarrow$\\
\midrule
WPs    & 3.21 &0.68\\
+Path     & \cellcolor{lgray} 4.49 &0.0 \\
\bottomrule
\end{tabular}
\end{center}}\end{minipage}
}
\subfloat[
\textbf{Vision encoder}.
\label{tab:encoder}
]{
\begin{minipage}{0.32\linewidth}{\begin{center}
\begin{tabular}{l|r}
\toprule
\textbf{}
& \multicolumn{1}{c}{\textbf{DS} $\uparrow$}\\
\midrule
Clip ViT & \cellcolor{lgray} 6.87  \\ %
w/o pretr. & 0.45 \\
Resnet-34 &  2.71 \\
\bottomrule
\end{tabular}
\end{center}}\end{minipage}
}
\subfloat[
\textbf{Early stopping}.
\label{tab:early}
]{
\begin{minipage}{0.30\linewidth}{\begin{center}
\begin{tabular}{l|r}
\toprule
\textbf{}
& \multicolumn{1}{c}{\textbf{DS} $\uparrow$}\\
\midrule
1300  &  3.93\\
1800 & 4.49 \\
2100 & \cellcolor{lgray} 6.87 \\
2400 & 6.35 \\
\bottomrule
\end{tabular}
\end{center}}\end{minipage}
}
\vspace{-0.2cm}
\caption{Ablations of different parts of \methodBase{}, showcasing the superiority of the disentangled output representation and the large impact of the correct threshold for early stopping. The score of the default configuration is highlighted in gray. All numbers are official Leaderboard 2.0 scores.}
\vspace{-0.2cm}
\label{tab:ablations}
\end{table}

\boldparagraph{Action Dreaming} \\
For the Action Dreaming evaluation, we use Success Rate as the metric. Each category has its own definition of success, which we detail in the following:
\begin{itemize}
\item \textbf{Slow down}: We calculate the target speed for each waypoint of the predicted speed waypoints. Those target speeds represent the target speeds for future timesteps. We do linear regression to get the slope. Success is defined as the slope being smaller than $-0.05 * v$, with $v$ being the current ego speed.
\item \textbf{Speed up}: Same calculation as for \textit{Slow down} but we use $slope > 0.05 * v$.
\item \textbf{Target Speed}: Since we do not know if the target speed can be reached in the prediction horizon of the waypoints we compare the predictions with the ground truth actions instead of directly comparing to the target speed. We use two rules defining success: First, if the predicted target speed inferred from the last two waypoints is in a 20\% range of the instructed target speed. Second, if the predicted target speed inferred from the last two waypoints is in the 20\% range of the speed of the last two waypoints of the ground truth speed waypoints. This can be different from the instructed target speed due to limitations in the acceleration rates of the vehicle.
\item \textbf{Lane Change}: We compare the final waypoint of the predicted path waypoint with the ground truth dreamer path and the ground truth expert path waypoints. We define the lane change as successful when the predicted final location is closer to the dreamer's final location than the expert final location.
\item \textbf{Objects (Collisions)}: This describes the task of driving towards or crashing into specific objects. We first look at the path. If the path of the expert trajectory and the ground truth dreamer trajectory is different (Average Displacement Error $ADE>1.0$) we count it as success if the predicted path is closer to the ground truth dreamer path than to the expert path ($ADE_{pred2expert} > ADE_{pred2dreamer}$).
If the dreamer path is nearly identical to the expert path ($ADE<1.0$) the instruction is about correct speed predictions (e.g., if the instruction is ``drive towards a dynamic object" it is important to get the speed right and not just the path. The success is then defined as $ADE_{pred2dreamer} < 1.0 $ and the average predicted speed is within 30\% of the ground truth dreamer speed.
\end{itemize}

\subsection{Quantitative Results}\label{sec:supp:quantitative}

\subsubsection{Ablations on CARLA Leaderboard 2.0}
We provide additional results on \methodBase{}.
\textit{Output representation.} Tab.~\ref{tab:output} compares the DS on the Leaderboard for the different output representations. As the goal of the additional path prediction is improved lateral control, we also report the collisions with static layout as this is mainly caused due to bad steering. With the disentangled representation, we can reduce the layout collision from 0.68 to 0 showing the strength of additional path predictions. \\
\textit{Vision-Language and CLIP pretraining.} We ablate the pretraining of the vision encoder and train the same model from scratch. Tab.~\ref{tab:encoder} 'w/o pretr.' shows that the pretraining stage is essential for good driving performance (longer training can further improve the performance but is unlikely to reach the performance of the pretrained model). Additionally, we show a comparison to the widely used Resnet-34 pretrained on ImageNet. The decreased performance (2.71 vs. 6.87 DS) indicates the importance of the larger ViT and the internet-scale image-language pretraining. \\
\textit{Early stopping.}
As described in the metric section the DS on long routes is not optimal and favors models that do not complete the full route. 
We ablate the thresholds for the early stopping as it is not trivial to calculate the perfect trade-off as the routes and density of scenarios are secret (however a rough function of the expected DS can be calculated, which we used to get a rough range). Tab.~\ref{tab:early} shows the Leaderboard DS for a given traveled distance in meters. This hyperparameter has a big impact on the final score. \\
\textit{Leaderboard variance.} We submitted our base model \methodBase{} with an early stopping threshold of 2100 and 2400 three times to the leaderboard to get an estimate of the evaluation variance. For the 2100 model, we obtain the following scores: 6.9, 5.5, and 5.3 resulting in a mean DS of 5.9 with a standard deviation of 0.87. The base model with a threshold of 2400 obtained 6.4, 6.3, and 4.8 resulting in a mean of 5.83 with a standard deviation of 0.90. 

\subsubsection{Bench2Drive Multi-Ability Results}
Tab.~\ref{tab:ability} shows the Bench2Drive Multi-Ability metrics. Consistent with the findings in the main paper \methodName{} outperforms existing works. Especially in the abilities \textit{Merging, Overtaking, Emergency Brake} and \textit{Traffic Sign} we get a large boost in performance. \textit{Give way} is still challenging.

\begin{table}[tb!]
\centering
\small
\scalebox{1.0}{\begin{tabular}{l|cc}
\toprule
 & \multicolumn{2}{c}{\textbf{Bench2Drive}} \\  
 & DS $\uparrow$ & SR(\%) $\uparrow$ \\ \midrule
w/o CoT & 84.41\scriptsize{$\pm$1.76} & 64.84\scriptsize{$\pm$2.42}\\
with CoT & \textbf{85.07\scriptsize{$\pm$0.95}} &	\textbf{67.27\scriptsize{$\pm$2.11}} \\
\bottomrule
\end{tabular}}
\vspace{-0.2cm}
\caption{\textbf{Inference Mode}. BEnch2Drive results for using \methodName{} with chain-of-thought (CoT) during inference and without. We see a small improvement when using CoT. 
\label{tab:cot}}
\vspace{-0.5cm}
\end{table}
\subsubsection{Chain-of-Thought Inference Mode}\label{sec:supp:cot}
We ablate the Chain-of-Thought (CoT) inference mode of \methodName{} in Tab.~\ref{tab:cot}. When using the Commentary task as CoT we see a small improvement so we decided to use this as the default mode. However, since without CoT the performance does not drop much, using the model without CoT is a feasible option especially when inference speed is important.

\subsection{Qualitative Results}\label{sec:supp:qualitative}
We provide more qualitative results of different navigational commands in Fig.~\ref{fig:nav}. Red dots are the predicted path waypoints and green are the predicted speed waypoints. Each row is captured from a closed-loop run while changing the navigation command. The second row shows how the model can successfully differentiate between different situations and adapt its behavior given a certain command. The vehicle starts in the right lane. When giving the instruction ``Turn left..." the model initiates a lane change to the left lane. After finishing this lane change it stays on this lane even though the command is still ``Turn left...". So the model learns to reason about the meaning of the different lanes, that it is forbidden to go on an oncoming lane, and that ``Turn left" not always mean to do a 90-degree turn. In the third row, we prompt the model with a misleading instruction: ``Turn right" on an intersection without a lane going to the right. The third and fourth image shows that the model is slightly confused but when unrolling closed-loop the car still goes straight and stays on the road. 
In the fourth row, we show out-of-distribution commands. The model is still able to choose a valid path when using a command that does not make sense like ``I really like my dog". When using the command ``Why is there a tree on the right side?" the model still picks the left turn, indicating that it does not just overfit on single words like \textit{left} or \textit{right} but also takes the context into account. The last image with the command ``Do a U-Turn" shows that the model is not capable of following concepts it has never seen during training. \\
Fig.~\ref{fig:dreamer_faster}, Fig.~\ref{fig:dreamer_slower}, Fig.~\ref{fig:dreamer_target}, Fig.~\ref{fig:dreamer_object} and Fig.~\ref{fig:dreamer_lane} shows qualitative results for the different Dreamer modes. In most cases, the model correctly follows the instructions even if it clearly goes against the expert driving behavior (e.g., accelerating at a red traffic light). We also include some of the failure cases of the model in the red boxes where the model ignores the instructions.

\newpage
\begin{figure*}[t]
\begin{center}
   \includegraphics[width=0.9\linewidth]{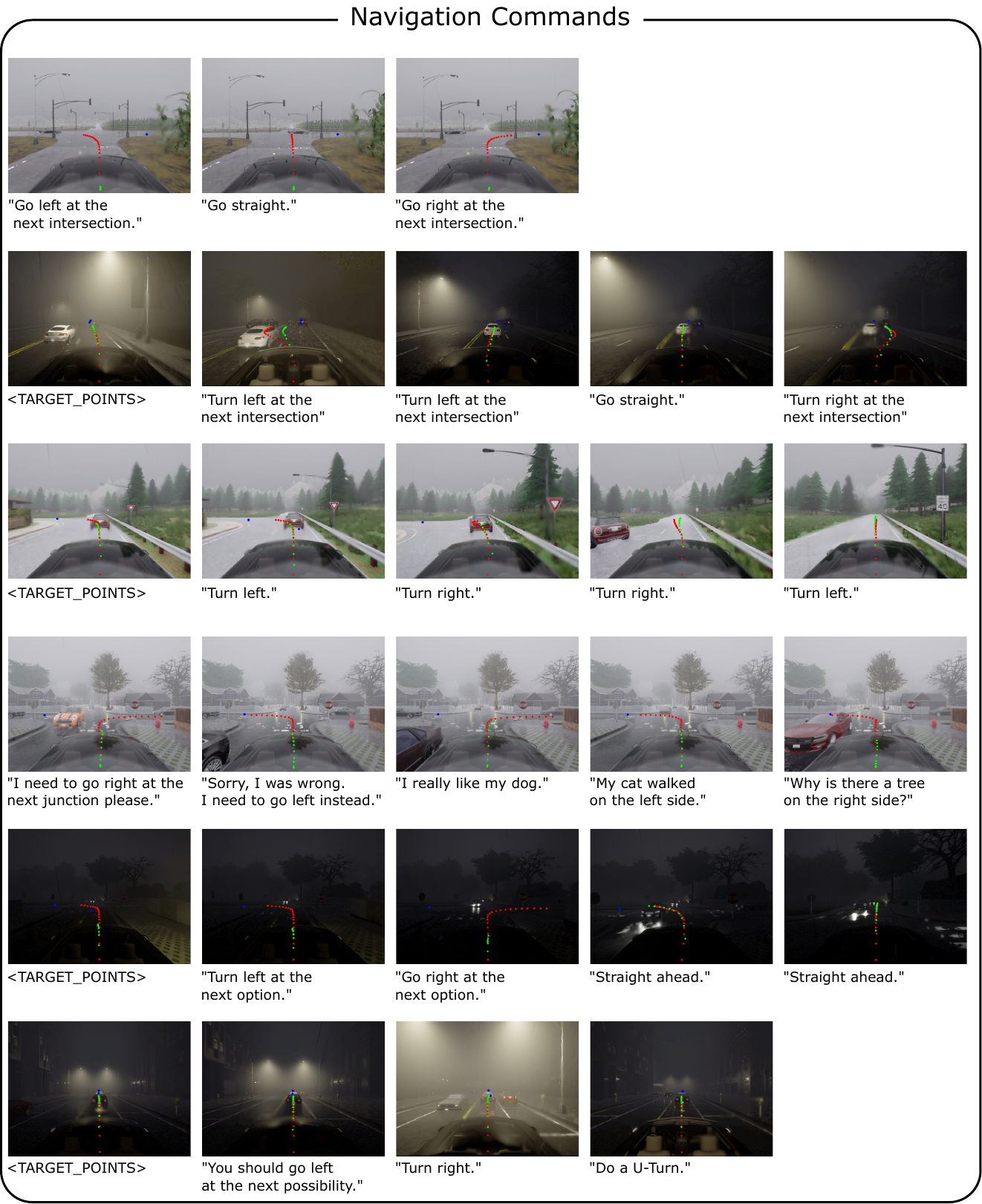}
\end{center}
\vspace{-0.4cm}
\caption{\textbf{Navigational Commands.} We show closed-loop results for different in-distribution and out-of-distribution commands. Red: path waypoints, green: speed waypoints.}
\label{fig:nav}
\end{figure*}

\newpage

\begin{figure*}[t]
\begin{center}
   \includegraphics[width=0.9\linewidth]{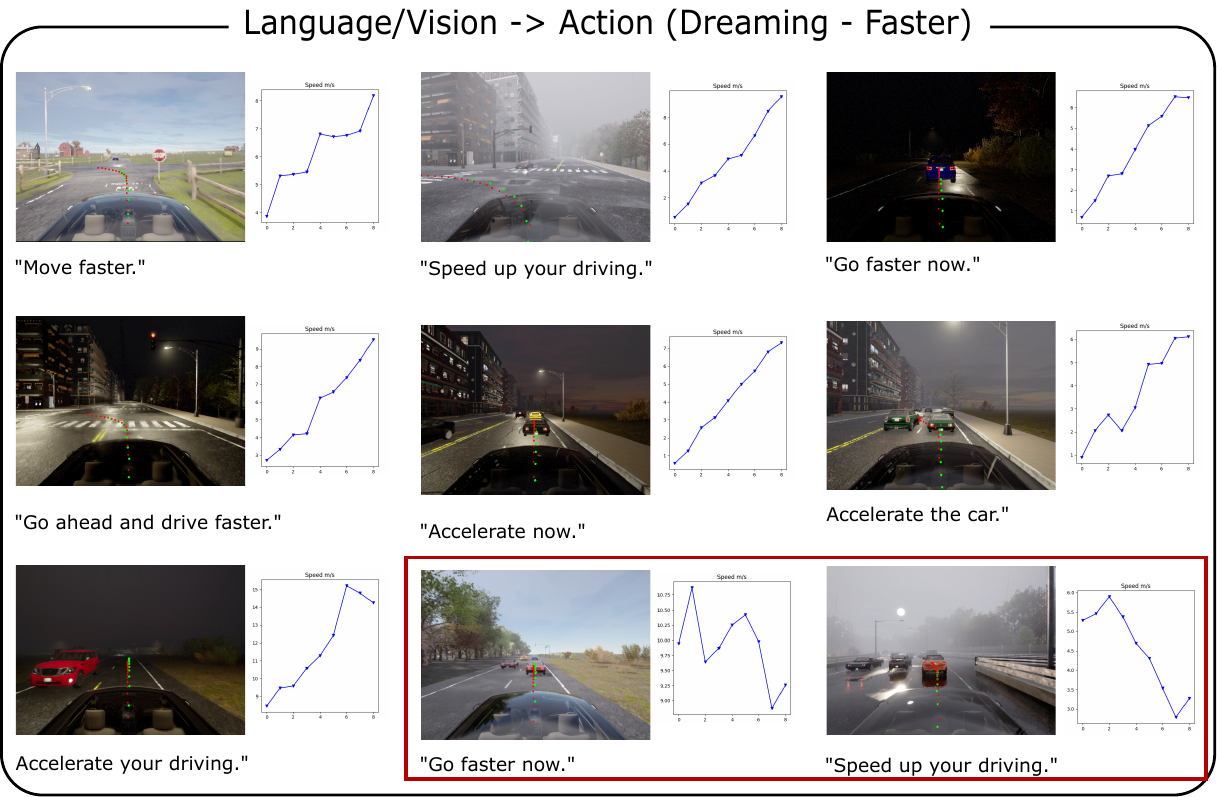}
\end{center}
\vspace{-0.4cm}
\caption{\textbf{Dreamer mode - Faster.} We show in blue the predicted speed curve (inferred from the speed waypoints). Inside the red border are examples  where the model does not follow the command correctly. Red: path waypoints, green: speed waypoints.}
\label{fig:dreamer_faster}
\end{figure*}

\begin{figure*}[t]
\begin{center}
   \includegraphics[width=0.9\linewidth]{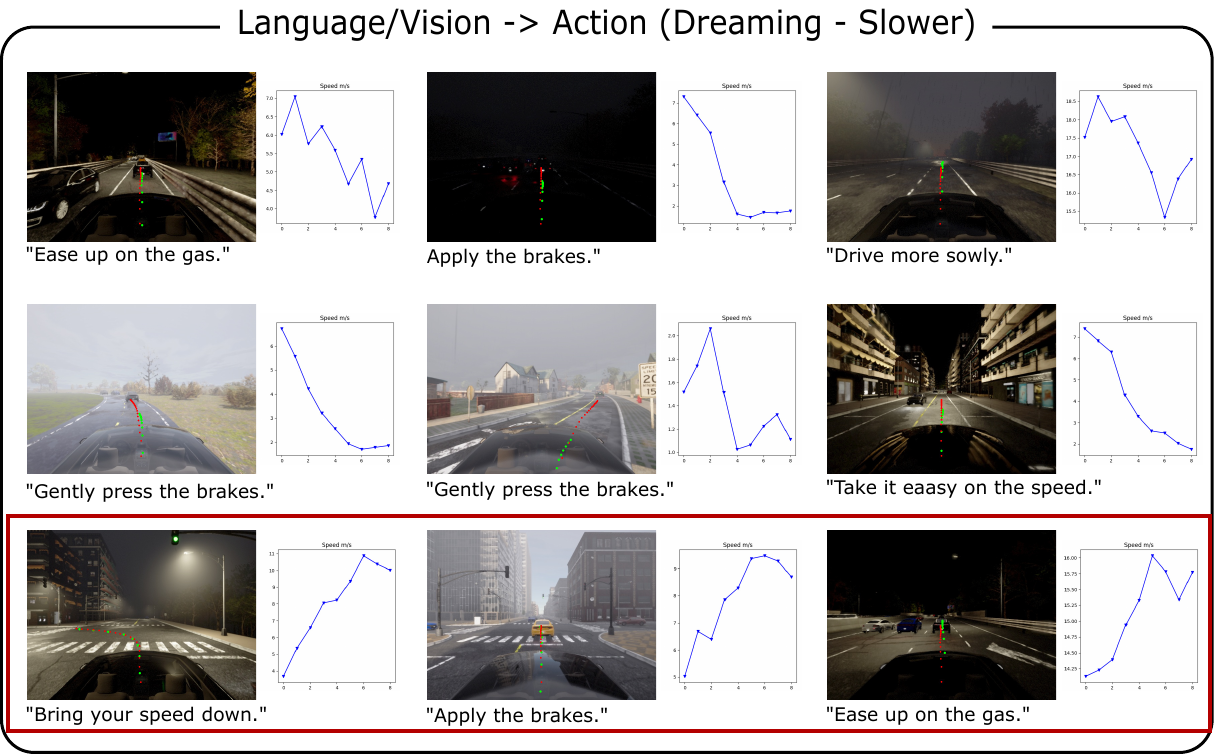}
\end{center}
\vspace{-0.4cm}
\caption{\textbf{Dreamer mode - Slower.} We show in blue the predicted speed curve (inferred from the speed waypoints). Inside the red border are examples where the model does not follow commands correctly. Red: path waypoints, green: speed waypoints.}
\label{fig:dreamer_slower}
\end{figure*}

\newpage

\begin{figure*}[t]
\begin{center}
   \includegraphics[width=0.9\linewidth]{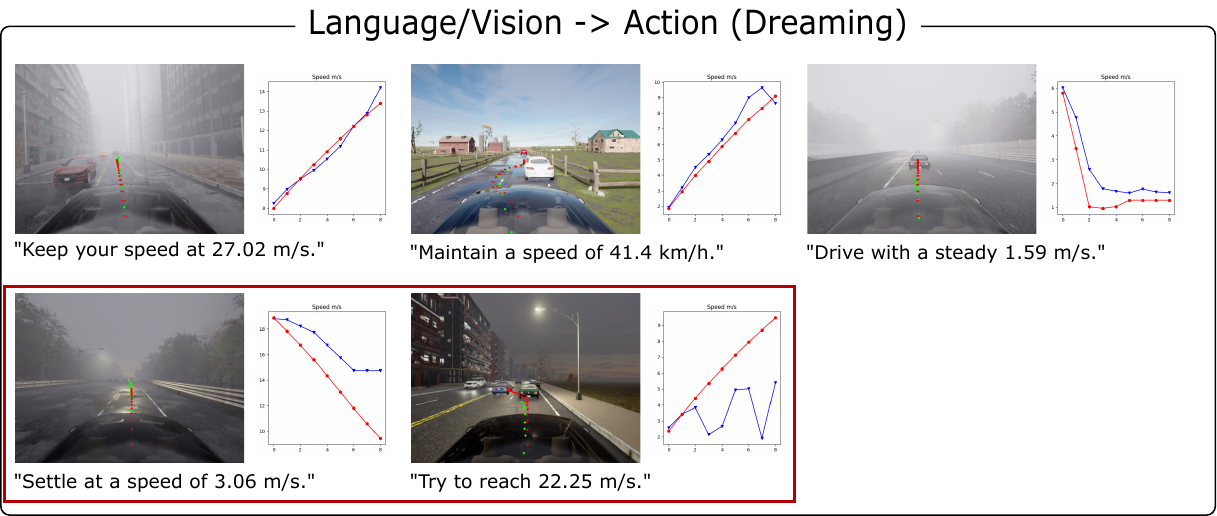}
\end{center}
\vspace{-0.4cm}
\caption{\textbf{Dreamer mode - Target speed.} We show in red line plot the ground truth speed curve and in blue the predicted one (inferred from the speed waypoints). Inside the red border are examples where the model does not follow commands correctly. Red: path waypoints, green: speed waypoints.}
\label{fig:dreamer_target}
\end{figure*}

\begin{figure*}[t]
\begin{center}
   \includegraphics[width=0.9\linewidth]{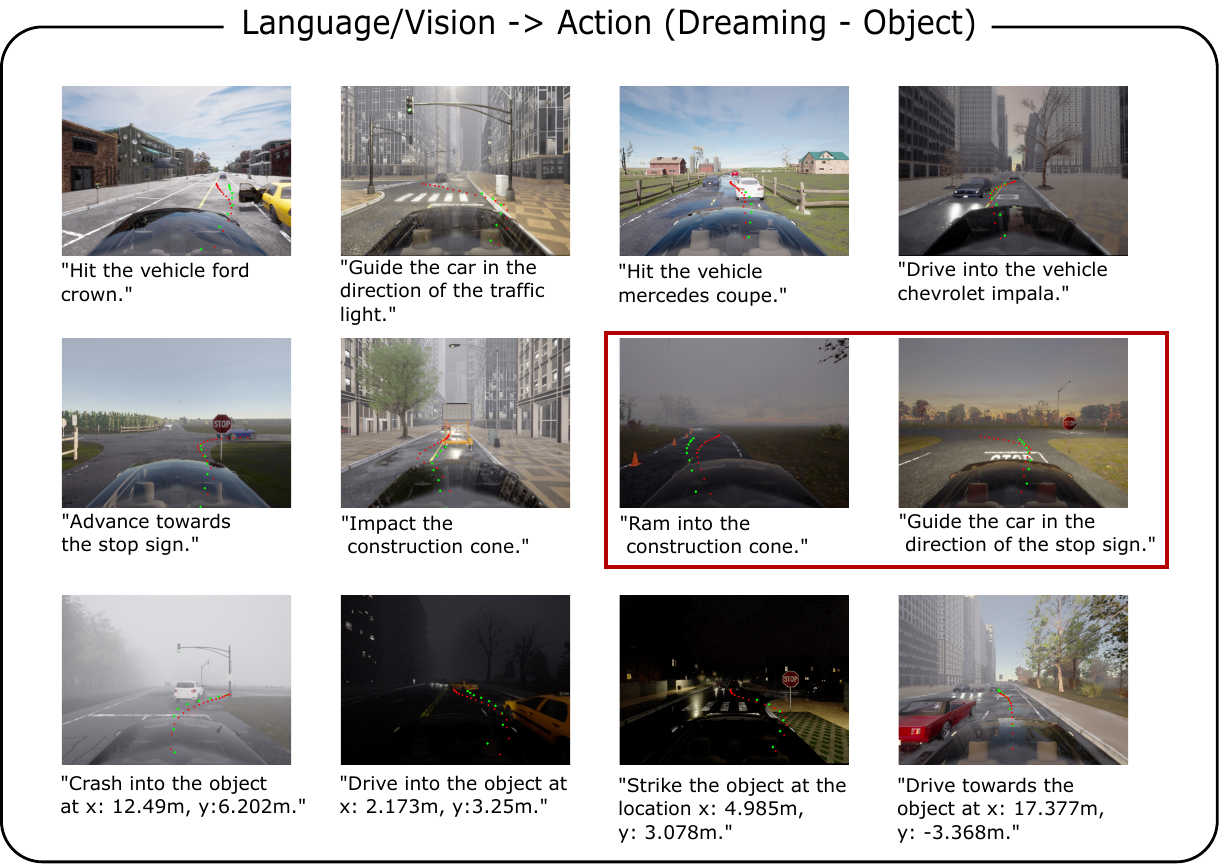}
\end{center}
\vspace{-0.4cm}
\caption{\textbf{Dreamer mode - Object (Collision).} Inside the red border are examples where the model does not follow dreamer mode command correctly. Red: path waypoints, green: speed waypoints.}
\label{fig:dreamer_object}
\end{figure*}

\begin{figure*}[t]
\begin{center}
   \includegraphics[width=0.9\linewidth]{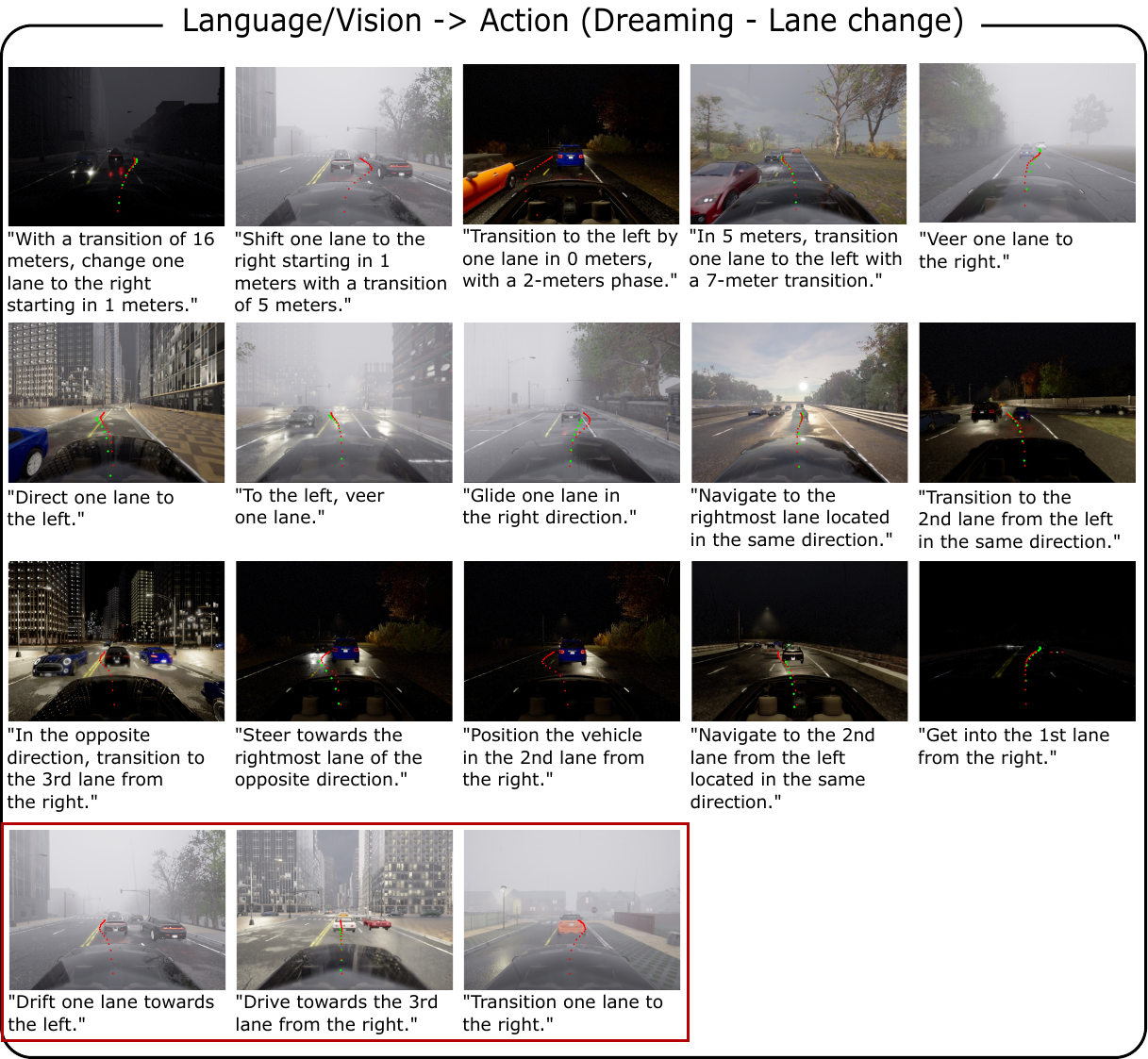}
\end{center}
\vspace{-0.4cm}
\caption{\textbf{Dreamer mode - lane changes.} Inside the red border are examples where the model does not follow dreamer mode commands correctly. Red: path waypoints, green: speed waypoints.}
\label{fig:dreamer_lane}
\end{figure*}

\end{document}